\documentclass[11pt]{article}

\usepackage[preprint]{acl}
\usepackage{times}
\usepackage{latexsym}
\usepackage[T1]{fontenc}
\usepackage[utf8]{inputenc}
\usepackage{microtype}
\usepackage{inconsolata}
\usepackage{graphicx}
\usepackage{booktabs}
\usepackage{multirow}
\usepackage{amsmath}
\usepackage{amssymb}
\usepackage{pgfplots} \pgfplotsset{compat=1.18}
\usepackage{tikz}
\usepackage{amssymb}
\usepackage{makecell}

\title{Where Experts Disagree, Models Fail:\\Detecting Implicit Legal Citations in French Court Decisions}
\author{Avrile Floro\textsuperscript{1} \\
  \texttt{avrile.floro@ip-paris.fr} \\\And
  Tamara Dhorasoo\textsuperscript{2} \\
  \texttt{dhorasoo.tamara@uphf.fr} \\\AND
  Soline Pellez\textsuperscript{2} \\
  \texttt{soline.pellez@uphf.fr} \\\And
  Nils Holzenberger\textsuperscript{1} \\
  \texttt{nils.holzenberger@telecom-paris.fr} \\\AND
  \\
  \textsuperscript{1}Télécom Paris, Institut Polytechnique de Paris
  \textsuperscript{2}Université Polytechnique Hauts-de-France
}
\begin{document}
\maketitle

\begin{abstract}
Applying computational methods to law at scale requires separating genuine legal reasoning from surface similarity. We study this through a concrete task: detecting \textit{implicit} citations of the French Civil Code, where a court applies a statutory rule without naming it (a \textit{post-hoc} question about the reasoning a court actually used). We release a benchmark of 1,015 passage--article pairs annotated by three legal experts. Our central finding is that their disagreement is itself informative: the third of cases the experts dispute are where models fail. Our best ensemble reaches an F1 score of 0.70 overall. Yet, two-thirds of its false positives fall on those disputed cases, a concentration that holds across all ten models we evaluate. Expert disagreement thus signals uncertainty in the task and in its gold label, which aggregate metrics hide. This should not block useful tools, however: reframed as top-$k$ ranking with multi-model consensus, the same signals reach 76\% precision for the top-200 candidates of the benchmark without supervision.
\end{abstract}

\section{Introduction}
A lawyer researching case law faces an asymmetry: explicit citations are trivial to find through keyword search. However, implicit applications, where a court applies a legal rule without naming it, are hidden. Consider a practitioner seeking examples of how article 2274 of the French Civil Code (the presumption of good faith) is applied in practice. Searching for ``article 2274'' retrieves decisions that explicitly cite this provision. Yet many decisions apply the same legal reasoning without numerical reference, using formulations such as ``the mere observation of the increase in rental debt is not sufficient to establish bad faith.'' This blind spot is also relevant to quantitative legal scholarship. Take a researcher studying whether French courts have expanded the scope of the good-faith presumption over the past decade. If the analysis captures only decisions that explicitly cite article 2274, it misses cases where the same provision is applied implicitly, and could potentially skew conclusions about jurisprudential trends. Our work begins to address this problem by evaluating the reliability of automatic detection. Specifically, this paper tackles the task of detecting \emph{implicit statutory citations}  (Figure~\ref{fig:explicit_implicit}): given a passage from a court decision and a candidate Civil Code article, determine whether the passage applies that article's legal rule without explicitly mentioning it. 

This task is both practically important and methodologically challenging. Indeed, it requires distinguishing genuine legal reasoning from semantic similarity. But how difficult is this task, and where do current methods fail?
We make four contributions\footnote{Data and code: \url{https://github.com/avrilemay/implicit-legal-citations} (code); \href{https://zenodo.org/records/21206799}{doi:10.5281/zenodo.21206799} (model \& data)} that characterize both the limits and the practical potential of computational approaches to this problem. 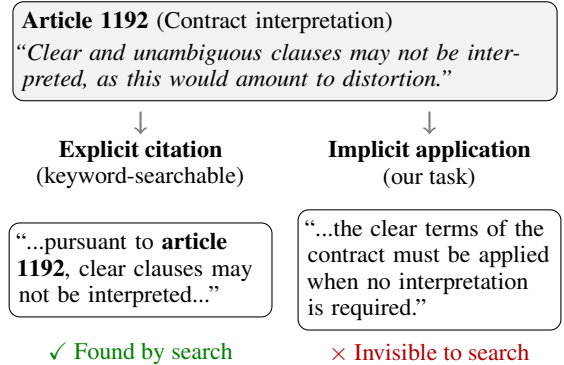
\begin{figure}[t]
\centering
\begin{tikzpicture}[
    box/.style={draw, rounded corners, text width=3.2cm, align=left, font=\footnotesize},
    artbox/.style={draw, rounded corners, text width=\columnwidth-0.8cm, align=left, font=\footnotesize, fill=gray!10},
    label/.style={font=\footnotesize\bfseries, align=center}
]
\node[artbox] (article) at (0, 0) {
    \textbf{Article 1192} (Contract interpretation)\\[2pt]
    \textit{``Clear and unambiguous clauses may not be interpreted, as this would amount to distortion.''}
};

\node[label] (explabel) at (-1.9, -1.5) {Explicit citation\\{\footnotesize\normalfont (keyword-searchable)}};
\node[label] (implabel) at (1.9, -1.5) {Implicit application\\{\footnotesize\normalfont (our task)}};

\node[box] (explicit) at (-1.9, -2.9) {
    ``...pursuant to \textbf{article 1192}, clear clauses may not be interpreted...''
};

\node[box] (implicit) at (1.9, -2.9) {
    ``...the clear terms of the contract must be applied when no interpretation is required.''
};

\node[font=\footnotesize, green!50!black] at (-1.9, -4.0) {$\checkmark$ Found by search};
\node[font=\footnotesize, red!70!black] at (1.9, -4.0) {$\times$ Invisible to search};

\draw[->, thick, gray] (-1.9, -0.8) -- (-1.9, -1.1);
\draw[->, thick, gray] (1.9, -0.8) -- (1.9, -1.1);
\end{tikzpicture}
\caption{Explicit and implicit statutory citations. While both excerpts apply article~1192, only the left one can be found by keyword search.}
\label{fig:explicit_implicit}
\vspace{-\baselineskip}
\end{figure} \textbf{First}, we introduce an adversarial benchmark for implicit citation detection in French civil law (\S\ref{sec:dataset}). We train a bi-encoder on explicit citations, use it to retrieve semantically similar candidates, then perform adversarial filtering using o3 with a conservative prompt. The final dataset comprises 1,015 pairs. \textbf{Second}, we conduct an annotation study with three legal experts (\S\ref{sec:annotation}). It reveals the difficulty of the task: experts frequently disagree on whether a passage applies legal reasoning or merely states facts. \textbf{Third}, we show that expert disagreement predicts model failure (\S\ref{sec:experiments}--\ref{sec:fp_analysis}). Our supervised ensemble achieves an overall F1 of 0.70.
This aggregate masks failure on disputed cases:
most of its false positives fall on the cases where annotators disagreed, a pattern observed across all ten models tested.
\textbf{Fourth}, despite these limits, we show a path toward practical assistance tools (\S\ref{sec:unsupervised}). Reframing classification as top-$k$ ranking and exploiting LLM consensus yields 76\% precision when $k$
is set to $200$ in an unsupervised setting.

\section{Related Work}

\paragraph{Legal NLP and Citation Detection} Explicit citation extraction identifies numerical statute references, and can be solved with straightforward pattern matching. Several tasks in legal NLP aim at connecting legal statutes and cases, where no explicit link exists between the two. Statutory reasoning is the task of predicting whether a given statute applies to a given case~\citep{holzenberger_dataset}. Legal Statute Identification is the more general task of deciding which statutory provisions are relevant to a given case from a closed set~\citep{paul_legal, paul2021lesicinheterogeneousgraphbasedapproach}. An open-set version of the task exists, as statute retrieval in the COLIEE competition~\citep{rabelo2024coliee}, with related prior-case-retrieval benchmarks~\citep{joshi-etal-2023-u, paul-etal-2025-il}. Our task differs from all three: instead of retrieving relevant articles for a factual query (an \textit{ex ante} prediction), we determine whether a court applies a specific article's legal rule within its reasoning (a \textit{post-hoc} detection). These two are independent: a decision may omit a statute that could apply and may, instead, invoke one that is of debatable relevance. The latter is precisely a case of interest to legal scholarship. Our task also differs from natural language inference because our negative categories (factual descriptions, party claims, special regimes) are often logically compatible with the article. Hence, an NLI model may predict entailment even though they do not involve any judicial application. \citet{havaldar-etal-2025-entailed} show that NLI models struggle to recognize entailments that are implied rather than explicit.


\paragraph{Legal-domain Language Models} Language models fine-tuned on legal-domain documents have generally improved performance on legal tasks~\citep{dominguezolmedo2024lawma}:
LegalBERT~\citep{chalkidis-etal-2020-legal} showed gains from domain pretraining;
JuriBERT~\citep{douka2021juribert}, which we use as our base encoder, achieves good results on French legal benchmarks. We further use LLMs fine-tuned through continued learning on legal documents, such as SaulLM~\citep{colombo2024saullm} and LawMA~\citep{dominguezolmedo2024lawma}. Legal LLMs must however reach a certain size to compete with larger, proprietary models~\citep{saullm54b}. Our experiments show that specialized models capture legal semantics better than general encoders. They are nonetheless vulnerable to failure on ambiguous cases. The model used in \S\ref{sec:dataset}, o3, ranks 12th on LegalBench~\citep{legalbench}, making it a reasonable baseline.\footnote{See \url{https://www.vals.ai/benchmarks/legal_bench}, consulted on February 19, 2026.}

\paragraph{Annotation Disagreement in NLP} It is the norm in NLP to consider annotator disagreement as noise and the majority vote is used to suppress it. However, recent work challenges this view. \citet{basile2021toward} and \citet{uma2021learning} argue for treating annotator variation as informative signal rather than noise. \citet{kim2025learning} distinguish ``consensual'' from ``non-consensual'' instances, showing that annotator agreement is predictive of model behavior. Our work provides a legal-domain example of this insight. While cases need to be decided upon, which justifies the use of a gold label, we have quantified the concentration of errors on disputed cases and demonstrated that this pattern is consistent across architectures. 
The boundary between fact and law has long been recognized as difficult to draw. \citet{savelka2018segmenting} find low agreement on distinguishing factual from legal sentences in U.S.\ court opinions.
The concept of legal ``open texture'' \citep{hart1961concept} offers an analogy: when a general rule is applied to particular facts, legal concepts have a core of settled meaning but a penumbra of uncertainty. In a clear case, the court simply states that the rule applies. In a borderline case, the court must first examine the facts in detail to decide whether it does. The sentence that carries this decision then reads like a finding of fact. This is why fact and law become hard to tell apart in such passages.


\section{Dataset Construction}
\label{sec:dataset}
We introduce a benchmark for detecting \emph{implicit} citations of the French Civil Code in court decisions, when a legal rule is applied with no numerical reference to the article.

\subsection{Source Data}
We used Judilibre, the open API of the French Ministry of Justice, as the main source for our data. Personal names are pseudonymized at the source. We restrict our scope to decisions from \emph{tribunaux judiciaires} (first-instance civil courts). At the time of collection in July 2025, only 86 of the 164 courts had published decisions and no data was available prior to December 2023. We collected 182,155 decisions with usable motivation sections (i.e., Judilibre's zoning delimits a non-empty, unredacted \textit{motivations} zone), spanning December 2023 to July 2025. We divided each motivation into chunks with a sentence-aware algorithm (Appendix~\ref{app:biencoder}). We focus on first-instance courts because they publish more decisions than appellate courts, broadening the coverage. Furthermore, in French civil law, only first-instance and appellate courts (not the Cour de cassation) can rule and interpret contracts. They are governed by the obligations articles that dominate the benchmark (Book III, 77\%, Table \ref{tab:book_distribution}). Finally, limiting the corpus to a single jurisdiction level ensures homogeneity.

\subsection{Bi-Encoder Training on Explicit Citations}
\label{subsec:biencoder}

We first train a retrieval model on \emph{explicit} citations, i.e. cases where chunks contain verbatim article numbers, extracted with regex patterns and TF-IDF filtering (details in Appendix~\ref{app:biencoder}). Before training, we mask all legal references in the text (article numbers, code names, statute references) with special tokens. We force the model to learn semantic associations rather than pattern matching. We fine-tune a JuriBERT-based bi-encoder with Multiple Negatives Ranking Loss \citep{henderson2017efficient}, encode all Civil Code articles and use a FAISS index \citep{johnson2019billion} for nearest-neighbor retrieval.

\subsection{Implicit Candidate Generation}\label{subsec:candidates}

Our objective is to uncover \emph{implicit} citations. We apply the bi-encoder to the held-out split of its data (Appendix~\ref{app:biencoder}). First, we eliminate chunks containing explicit keywords such as `article', `loi', `code'.
Second, for each chunk, we retrieve the top-5 nearest articles. Finally, a candidate is discarded if its article is cited elsewhere in the decision. We also account for statutory renumbering (Appendix~\ref{app:renumbering}). This yields 40,566 candidate pairs.

\subsection{Adversarial Filtering with o3}\label{subsec:filtering} 

We applied adversarial filtering \citep{zellers-etal-2018-swag} to select challenging examples. We keep candidates labeled as positive by a strong model. The 565 negatives identified after human annotation are cases where the model failed. This creates difficult negative cases, where models sharing similar reasoning patterns with the one being used as a filter will likely fail. We used OpenAI's o3 model to evaluate these pairs. The prompt is engineered to be strict since the model is asked to reply `NO' in case of doubt (full prompt in Appendix~\ref{app:o3prompt}). From this pool, o3 returned 4,206 positive predictions (10.4\%), involving 497 articles. In accordance with the adversarial filtering objective, we only selected the pairs where o3 predicted the implicit use of an article. Finally, to avoid circularity, o3 was used only to create the dataset and was not evaluated afterwards on the final task. 

\subsection{Selection for Annotation} \label{subsec:selection}

From the 4,206 pairs predicted positive by o3, one of the legally-trained authors selected pairs for annotation, prioritizing coverage across distinct Civil Code articles.
The final dataset contains 1,015 pairs of (chunk, article) over 418 Civil Code articles (per-article frequency in Appendix~\ref{app:datasetstats}). Table~\ref{tab:book_distribution} shows the distribution across the books of the Civil Code. Book III covers contracts, successions, and matrimonial regimes. It accounts for 77\% of the dataset, which is expected because cases related to obligations dominate first-instance civil litigation. Length statistics are reported in Appendix~\ref{app:datasetstats}.

\begin{table}[h]
\centering
\small
\begin{tabular}{lrl}
\toprule
\textbf{Book} & \textbf{Art.} & \textbf{Pairs} \\
\midrule
I --- Persons       &  52 &  \phantom{0}86 (8.5\%) \\
II --- Property    &  11 &  \phantom{0}21 (2.1\%) \\
III --- Obligations & 310 & 785 (77.3\%) \\
IV --- Security     &  43 & 121 (11.9\%) \\
V --- Mayotte       &   \phantom{00}2 &   \phantom{00}2 (0.2\%) \\
\bottomrule
\end{tabular}
\caption{Dataset distribution across Civil Code books.}
\label{tab:book_distribution}
\vspace{-\baselineskip}
\end{table}

\subsection{Bounding Selection Bias} \label{par:bounding-selection-bias}
Every retained pair was predicted positive by o3. Because the benchmark is a filtered sample, it is possible that o3 wrongly rejected genuine implicit citations. To bound this risk, we drew a control sample of 100 (chunk, article) pairs retrieved by the bi-encoder but \textit{rejected} by o3, stratified over 100 distinct Civil Code articles to avoid article-frequency confounds. They were labeled by the same three legal experts under the protocol of \S\ref{sec:annotation} (A\textsubscript{1} and A\textsubscript{2} independently, A\textsubscript{3} adjudicating disagreements). A\textsubscript{1} judged all 100 non-implicit and A\textsubscript{2} judged 98 non-implicit. A\textsubscript{3} resolved the two disputed pairs as non-implicit (facts only). No pair was confirmed as an implicit application. With zero positives in 100 trials, the rate of missed implicit citations among o3-rejected candidates is bounded below roughly 3\% (95\%~CI). This is an upper bound on the selection bias, not an evidence of its absence.

\section{Annotation Study}
\label{sec:annotation}

\paragraph{Protocol}

Annotation was carried out independently by three of the authors, all legally trained in French civil law. Each entry pair was reviewed and assigned a label: YES, when the article was deemed to be applied implicitly, or NO otherwise. Negative cases could be further qualified (e.g., the passage states facts, or relies on a different legal regime). Appendix~\ref{app:interface} details the annotation interface built for the campaign. Annotators A\textsubscript{1} and A\textsubscript{2} labeled the whole dataset. A\textsubscript{3} was the adjudicator for the 339 cases where A\textsubscript{1} and A\textsubscript{2} disagreed.

\paragraph{Agreement Statistics}

Agreement between A\textsubscript{1} and A\textsubscript{2} reached 66.6\% (Cohen's ${\kappa = 0.33}$), consistent with prior work on implicit reasoning tasks \citep{savelka2018segmenting, troiano2019crowdsourcing}. A\textsubscript{3} resolved the 339 disputed cases (33.4\%), for a total of 450 YES and 565 NO. Table~\ref{tab:confusion_annotators_main} shows that A\textsubscript{2} labeled 67\% of cases as YES versus 50\% for A\textsubscript{1}. This asymmetry explains the moderate $\kappa$. Across the disputed cases, A\textsubscript{3} sided with A\textsubscript{1} in 73.5\% of cases, versus 26.5\% with A\textsubscript{2}. This suggests that A\textsubscript{1} adopted a more restrictive reading.

\begin{table}[h]
\centering
\begin{tabular}{lcc}
\toprule
& \textbf{A\textsubscript{2}=Yes} & \textbf{A\textsubscript{2}=No} \\
\midrule
\textbf{A\textsubscript{1}=Yes} & 425 & \phantom{0}83 \\
\textbf{A\textsubscript{1}=No} & 256 & 251 \\
\bottomrule
\end{tabular}
\caption{Confusion matrix between primary annotators.}
\label{tab:confusion_annotators_main}
\end{table}

\paragraph{Structure of Disagreement}

The decisions of Annotator A\textsubscript{3} on the 339 disputed cases
reveal two patterns (Table~\ref{tab:disagreement_structure} in Appendix~\ref{app:disagreement}). 92.6\% of disagreements resolve to NO: borderline cases are more often judged as non-implicit in the end. Moreover, the fact versus law boundary dominates in the disputed cases. 43.4\% of disagreements involve chunks describing facts or party claims without judicial reasoning, according to A\textsubscript{3}'s categorization. This pattern is consistent with the distinction drawn by \citet{hart1961concept} between the ``core'' of legal concepts (clear cases) and their ``penumbra'' (borderline cases where application is genuinely uncertain). 

\paragraph{Implications for Evaluation}

Recent works state that annotator disagreement should not be interpreted  as mere noise, but as a signal reflecting genuine ambiguity \citep{aroyo2015truth, pavlick2019inherent, uma2021learning}. We embrace this perspective: the 339 disputed cases may constitute instances where trained experts reached different conclusions. This prompts our question: do models fail indiscriminately, or do their errors concentrate upon these disputed cases?

\section{Experiments}
\label{sec:experiments}
We evaluated several approaches on the task of implicit citation detection. Although we frame annotations as binary classification, the ultimate goal is retrieval: given a passage from a Court decision, identify which Civil Code articles are implicitly applied, if any.
We therefore progress from supervised classification (§\ref{sec:supervised}) to zero-shot LLM classification (§\ref{sec:zeroshot}), and finally to unsupervised ranking that combines multiple signals (§\ref{sec:unsupervised}).

\subsection{Supervised Classifiers}
\label{sec:supervised}

\paragraph{Models} We evaluate several pre-trained encoders on the classification task. To avoid overfitting considering the limited size of our dataset (1,015 entries), we freeze the encoders' parameters, extract sentence-level representations, and then train simple classifiers on top. We test eight encoders across three categories: general French models CamemBERT \citep{martin-etal-2020-camembert} and CamemBERTav2 \citep{antoun2024camembert2}; legal-domain models JuriBERT \citep{douka2021juribert}, SaulLM-7B \citep{colombo2024saullm}, and Lawma \citep{dominguezolmedo2024lawma}; and general-purpose models LLaMA-3.1-8B \citep{grattafiori2024llama3}, MPNet \citep{song2020mpnet}, and MiniLM \citep{wang2020minilm}. All experiments rely on 5-fold cross-validation, with folds grouped by decision ID. Each fold contains 203 examples, for a total of 1,015. Performance is primarily assessed using Matthews Correlation Coefficient (MCC), which is more informative than raw accuracy under class imbalance. For each encoder, we vary input formatting, pooling methods, layer selection, and the classifier head. Additional details are reported in Appendix~\ref{app:gridsearch}. As a baseline, we use TF–IDF with logistic regression.

\begin{table}[h]
\centering
\small
\begin{tabular}{llccc}
\toprule
\textbf{Model} & \textbf{Head} & \textbf{F1} & \textbf{Acc} & \textbf{MCC} \\
\midrule
SAUL & LR & .69 & .74 & .47 \\
LLaMA & MLP1 & .69 & .74 & .46 \\
LawMA & MLP1 & .64 & .73 & .46 \\
ST-MPNet & MLP2 & .63 & .72 & .43 \\
CamemBERT & MLP2 & .67 & .72 & .43 \\
CamemBERTav2 & MLP1 & .65 & .71 & .42 \\
TF-IDF & LR & .63 & .71 & .41 \\
JuriBERT & LR & .62 & .69 & .37 \\
ST-MiniLM & MLP2 & .61 & .69 & .37 \\
\bottomrule
\end{tabular}
\caption{Supervised classification results (5-fold CV, F1 on positive class). Configuration details in Appendix~\ref{app:gridsearch}.}
\label{tab:supervised}
\end{table}

Table~\ref{tab:supervised} shows that among individual models, the strongest result is obtained by SAUL-7B with an MCC of 0.47. The TF–IDF baseline is competitive (\mbox{MCC = 0.41}) and surpasses JuriBERT (\mbox{MCC = 0.37}), despite the latter being pre-trained on French legal corpora. It suggests that lexical overlap between articles and judicial passages captures part of the signal for this task.

\paragraph{Ensemble}
To improve performance, we move beyond single models and construct ensembles with nested cross-validation. We explore weighted averaging, stacking, and rank-based fusion. The highest-scoring configurations all achieve the same MCC of 0.53 (Appendix~\ref{app:ensemble}). We retain the stacking
ensemble with a logistic-regression meta-learner combining CamemBERTav2, JuriBERT, LLaMA-3.1-8B, and SAUL-7B. It reaches an F1 of 70\% on the positive class and an accuracy of 77\%. Both legal-domain models (JuriBERT for French, SAUL-7B for English) appear in every top-performing ensemble, suggesting their representations complement those of generalist encoders. Appendix~\ref{app:ensemble} houses the full results.

\paragraph{Purpose-Built Entailment and Cross-Encoder Baselines} To test whether the difficulty is specific to the frozen-encoder models, we add two models that are architecturally distinct and fine-tuned: XLM-RoBERTa-large fine-tuned on XNLI (an encoder purpose-built for entailment) and a fine-tuned BGE-reranker-v2-m3 cross-encoder (same 5-fold CV, Appendices \ref{app:bge} and \ref{app:nli}). Neither matches the supervised ensemble. More importantly, both reproduce the disagreement-failure pattern of \S\ref{sec:fp_analysis}: their false positives concentrate on disputed true negatives (OR = $2.26$ for XLM and $2.22$ for BGE, both $p=0.0001$), within the range of  Figure~\ref{fig:fpr_by_agreement}. The concentration is therefore not an artifact of the models we use.

\subsection{Zero-Shot LLM Approaches}
\label{sec:zeroshot}
Our dataset was constructed using o3 as an adversarial filter (\S\ref{sec:dataset}). It raises the question of whether other LLMs share similar reasoning patterns. We evaluate ten instruction-tuned LLMs with zero-shot prompting (Appendix~\ref{app:prompt}). We include models from the Qwen family
\citep{qwen2024qwen25, qwen2025qwen3} (2.5-7B, 2.5-32B, and
3-32B with thinking mode), LLaMA-3.1
\citep{grattafiori2024llama3} (8B and 70B), as well as
Mistral-Nemo-12B \citep{mistral2024nemo}, Gemma-2-27B
\citep{gemma2024gemma2}, Aya-Expanse-32B \citep{dang2024aya},
Command-R-35B \citep{cohere2024commandr}, and the legal-domain
SaulLM-7B \citep{colombo2024saullm}. Model sizes range from 7B
to 70B parameters.

\paragraph{Results} Table~\ref{tab:zeroshot} reports performance (full results per annotator in Appendix~\ref{app:zeroshot}). Most models over-predict positive cases, with yes-rates ranging from 22\% to 96\%, far from the 44\% base rate. This positive bias validates our adversarial construction: most off-the-shelf models have similar accuracy as o3. Curiously, despite SAUL-7B achieving the best supervised performance (\mbox{MCC = 0.47}), it shows extreme positive bias in a zero-shot setting (95\% yes-rate, \mbox{MCC = 0.03}). It suggests that legal pre-training improves representation quality but does not calibrate the model's zero-shot judgment. This agrees with findings showing that the benefits of domain-specific pre-training
depend on the task \citep{gururangan2020dont}. Retrieval-augmented few-shot prompting on the three zero-shot models with the best performance (Qwen-2.5-32B, Qwen-2.5-7B, and 
LLaMA-3.1-8B) does not approach the ensemble either (Appendix~\ref{app:fewshot}). The best gain is +0.02 MCC.

\begin{table}[h]
\centering
\small
\begin{tabular}{lrrrr}
\toprule
\textbf{Model} & \textbf{Yes\%} & \textbf{F1} & \textbf{Acc} & \textbf{MCC} \\
\midrule
Qwen-2.5-32B & 66 & .66 & .63 & \textbf{.31} \\
Qwen-2.5-7B & 54 & .63 & .63 & .28 \\
LLaMA-3.1-8B & 22 & .47 & .65 & .28 \\
LLaMA-3.1-70B$^\dagger$ & 84 & .65 & .55 & .25 \\
Mistral-Nemo-12B & 75 & .63 & .56 & .21 \\
Gemma-2-27B & 94 & .63 & .49 & .17 \\
Qwen3-32B$^\ddagger$ & 85 & .63 & .52 & .16 \\
Aya-Expanse-32B & 94 & .62 & .47 & .08 \\
Command-R-35B & 96 & .61 & .46 & .04 \\
SAUL-7B & 95 & .61 & .46 & .03 \\
\bottomrule
\end{tabular}
\caption{Zero-shot LLM results ($n=1{,}015$, F1 on positive class). $^\dagger$4-bit quantization. $^\ddagger$Reasoning model with thinking mode.}
\label{tab:zeroshot}
\vspace{-\baselineskip}
\end{table}

The positive bias that is shared across models makes zero-shot LLMs unreliable binary classifiers. However, the wide range of models' yes-rate can be leveraged into a new strategy. Instead of using the predictions of a single model, we can treat the agreement among differently calibrated models as a ranking signal.

\subsection{Toward Unsupervised Retrieval}
\label{sec:unsupervised}

We use four LLMs with the lowest positive prediction rates: LLaMA-3.1-8B, Qwen-2.5-7B, Qwen-2.5-32B, and Mistral-Nemo-12B (Table~\ref{tab:zeroshot}). Their yes-rates, though still significant, are lower than those of the remaining models. We hypothesize that overly permissive models flood the ranking with weak candidates, which is undesirable in this setting. We use various ranking strategies. \textbf{Lexical baselines} uses TF-IDF and BM25 similarity between chunk and article. \textbf{Cross-encoder} refers to BGE-reranker-v2-m3 \citep{xiao2024bge} used out-of-the-box. It is a multilingual encoder that is applied directly to the French chunk-article pairs.
\textbf{LLMs intersection} ranks candidates by agreement among LLMs:  \textsc{Union} (at least one positive), \textsc{Inter2} (2 positives), \textsc{Inter3}, and \textsc{Inter4} (full agreement). Ties are broken using the cross-encoder score. \textbf{Unsupervised Ensemble} combines the four selected LLM predictions with equal weights and gives a bonus for agreement. We also include the normalized scores from \textsc{TF--IDF}, \textsc{BM25}, and the cross-encoder. The weights were set heuristically without tuning on labels to remain truly unsupervised. The ranking is stable across weight choices. The formula and sensitivity study are provided in Appendix \ref{app:ablation}.

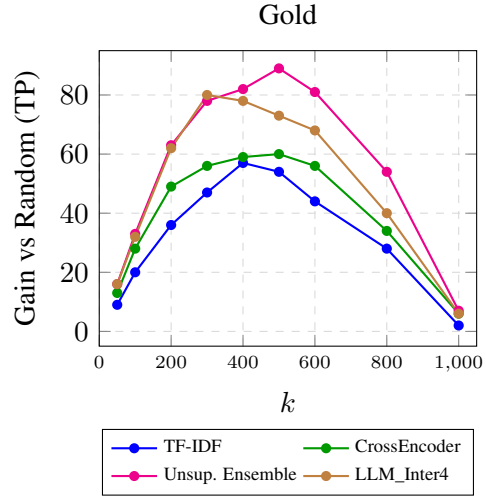
\begin{figure}[h]
\centering
\begin{tikzpicture}
\begin{axis}[
    width=0.85\columnwidth,
    height=5.5cm,
    xlabel={$k$},
    ylabel={Gain vs Random (TP)},
    xmin=0, xmax=1050,
    ymin=-5, ymax=95,
    xtick={0,200,400,600,800,1000},
    xticklabel style={font=\scriptsize},
    legend style={
        font=\scriptsize,
        at={(0.5,-0.28)},
        anchor=north,
        legend columns=2,
        cells={anchor=west},
    },
    grid=major,
    grid style={dashed,gray!30},
    title={Gold}
    ]
\addplot[blue, mark=*, thick, mark size=1.5pt] coordinates {
    (50,9) (100,20) (200,36) (300,47) (400,57) (500,54) (600,44) (800,28) (1000,2)
};
\addlegendentry{TF-IDF}
\addplot[green!60!black, mark=*, thick, mark size=1.5pt] coordinates {
    (50,13) (100,28) (200,49) (300,56) (400,59) (500,60) (600,56) (800,34) (1000,6)
};
\addlegendentry{CrossEncoder}
\addplot[magenta, mark=*, thick, mark size=1.5pt] coordinates {
    (50,16) (100,33) (200,63) (300,78) (400,82) (500,89) (600,81) (800,54) (1000,7)
};
\addlegendentry{Unsup. Ensemble}
\addplot[brown, mark=*, thick, mark size=1.5pt] coordinates {
    (50,16) (100,32) (200,62) (300,80) (400,78) (500,73) (600,68) (800,40) (1000,6)
};
\addlegendentry{LLM\_Inter4}
\end{axis}
\end{tikzpicture}
\caption{Gain in true positives retrieved compared to random ranking on gold labels.}
\label{fig:gain_vs_random}
\end{figure}

\paragraph{Results} Table~\ref{tab:ranking_ap} reports Average Precision (AP). The zero-shot ensemble reaches an AP of 0.67, above the random ranking by +0.23. The \textsc{Inter4} intersection method also performs well, reaching \mbox{AP = 0.66}. The consensus acts as a strong signal, even in an unsupervised environment. Table~\ref{tab:ranking_practical} shows precision and recall at various cutoffs. At ${k=200}$, precision is 76\% with 34\% recall. Figure~\ref{fig:gain_vs_random} confirms the unsupervised ensemble outperforms baselines  (full results per annotator in Appendix~\ref{app:ranking}). We discuss practical implications in \S\ref{sec:discussion}.

\begin{table}[h]
\centering
\small
\begin{tabular}{lcc}
\toprule
\textbf{Method} & \textbf{AP} & \textbf{$\Delta$ vs Random} \\
\midrule
Random & .44 & -- \\
TF-IDF & .57 & +.13 \\
BM25 & .57 & +.13 \\
Cross-Encoder & .61 & +.17 \\
\midrule
LLM\_Union & .62 & +.18 \\
LLM\_Inter2 & .63 & +.19 \\
LLM\_Inter3 & .65 & +.20 \\
LLM\_Inter4 & .66 & +.21 \\
\midrule
Unsup. Ensemble & \textbf{.67} & \textbf{+.23} \\
\bottomrule
\end{tabular}
\caption{Average Precision for unsupervised ranking methods on gold labels ($n=1{,}015$).}
\label{tab:ranking_ap}
\end{table}

\begin{table}[h]
\centering
\small
\begin{tabular}{lrrrrr}
\toprule
$k$ & \textbf{TP} & \textbf{FP} & \textbf{P@$k$} & \textbf{R@$k$} & \textbf{FP red. (\%)} \\
\midrule
50 & \phantom{0}38 & 12 & .76 & .08 & 57 \\
100 & \phantom{0}77 & 23 & .77 & .17 & 59 \\
200 & 151 & 49 & .76 & .34 & 56 \\
300 & 211 & 89 & .70 & .47 & 47 \\
\bottomrule
\end{tabular}
\caption{Unsupervised ensemble ranking at various cutoffs. FP red.\ = reduction vs.\ random.}
\label{tab:ranking_practical}
\vspace{-\baselineskip}
\end{table}

\paragraph{Limits} The same difficulty observed with supervised classifiers persists with unsupervised ranking.
At ${k=100}$, 65\% of false positives (15/23) come from the 33\% of disputed cases (Table~\ref{tab:ranking_fp_main}, Appendix \ref{app:ranking}).
This concentration calls for a systematic analysis.

\section{Discussion}
\label{sec:discussion}

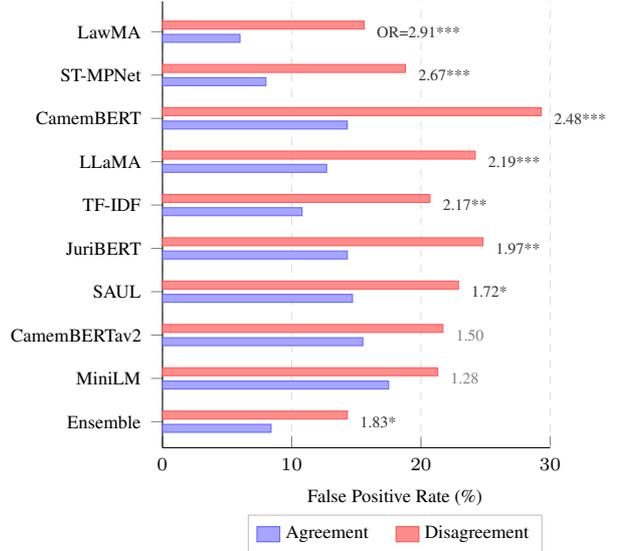
\begin{figure}[h]
\centering
\begin{tikzpicture}
\begin{axis}[
    width=0.9\columnwidth,
    height=7.5cm,
    xbar,
    bar width=3pt,
    xlabel={False Positive Rate (\%)},
    xmin=0, xmax=36,
    xtick={0,10,20,30},
    xticklabel style={font=\scriptsize},
    xlabel style={font=\scriptsize},
    ytick={0,1,2,3,4,5,6,7,8,9},
    yticklabels={Ensemble, MiniLM, CamemBERTav2, SAUL, JuriBERT, TF-IDF, LLaMA, CamemBERT, ST-MPNet, LawMA},
    yticklabel style={font=\scriptsize},
    ymin=-0.6, ymax=9.7,
    legend style={
        font=\scriptsize,
        at={(0.5,-0.15)},
        anchor=north,
        legend columns=2,
        draw=gray!50,
        /tikz/every even column/.append style={column sep=5pt},
    },
    legend image code/.code={
        \draw[#1, draw opacity=1] (0cm,-0.1cm) rectangle (0.3cm,0.1cm);
    },
    xmajorgrids=true,
    ymajorgrids=false,
    grid style={dashed, gray!25},
    clip=false,
    axis y line*=left,
    axis x line*=bottom,
]
\addplot[fill=blue!35, draw=blue!60, fill opacity=0.85] coordinates {
    (8.4,  0)
    (17.5, 1)
    (15.5, 2)
    (14.7, 3)
    (14.3, 4)
    (10.8, 5)
    (12.7, 6)
    (14.3, 7)
    (8.0,  8)
    (6.0,  9)
};
\addlegendentry{Agreement}
\addplot[fill=red!45, draw=red!65, fill opacity=0.85] coordinates {
    (14.3, 0)
    (21.3, 1)
    (21.7, 2)
    (22.9, 3)
    (24.8, 4)
    (20.7, 5)
    (24.2, 6)
    (29.3, 7)
    (18.8, 8)
    (15.6, 9)
};
\addlegendentry{Disagreement}
\node[font=\tiny, right, text=black!80] at (axis cs:15.6, 9) [xshift=1pt] {OR=2.91***};
\node[font=\tiny, right, text=black!80] at (axis cs:18.8, 8) [xshift=1pt] {2.67***};
\node[font=\tiny, right, text=black!80] at (axis cs:29.3, 7) [xshift=1pt] {2.48***};
\node[font=\tiny, right, text=black!80] at (axis cs:24.2, 6) [xshift=1pt] {2.19***};
\node[font=\tiny, right, text=black!80] at (axis cs:20.7, 5) [xshift=1pt] {2.17**};
\node[font=\tiny, right, text=black!80] at (axis cs:24.8, 4) [xshift=1pt] {1.97**};
\node[font=\tiny, right, text=black!80] at (axis cs:22.9, 3) [xshift=1pt] {1.72*};
\node[font=\tiny, right, text=gray]              at (axis cs:21.7, 2) [xshift=1pt] {1.50};
\node[font=\tiny, right, text=gray]              at (axis cs:21.3, 1) [xshift=1pt] {1.28};
\node[font=\tiny, right, text=black!80] at (axis cs:14.3, 0) [xshift=1pt] {1.83*};

\draw[white, line width=1.5pt] (axis cs:30,-0.6) -- (axis cs:36,-0.6);
\end{axis}
\end{tikzpicture}
\caption{False positive rates by annotator agreement status (agree versus disagree). Odds ratios right of bars. Significance after FDR correction: {***}\,$p<.001$, {**}\,$p<.01$, {*}\,$p<.05$; gray~=~n.s.}
\label{fig:fpr_by_agreement}
\end{figure}

\paragraph{False Positives Concentrate on Disagreement Cases} \label{sec:fp_analysis}
The previous sections found that legal experts disagree on implicit citation detection, and that both supervised and unsupervised methods reach reasonable performance when aggregated. We now examine the relationship between these two findings: do model errors distribute uniformly? Do they concentrate on the cases that were difficult for human annotators? Our supervised ensemble predicts 66 false positives. Table~\ref{tab:fp_distribution} shows that these errors concentrate on cases of annotator disagreement. The false positive rate is also higher when annotators disagree, by a factor of 1.7 (14.3\% vs.\ 8.4\%).  To ensure this pattern is robust, we examine false positive rates for the nine individual classifiers used in the supervised experiments. We focus on true negatives (gold = NO, $n=565$). We compare error rates on consensual negatives versus disagreed ones. After FDR correction, seven of the nine  models show significantly higher false positive rates on disagreement cases, as does the supervised ensemble. Figure~\ref{fig:fpr_by_agreement} shows that odds ratios (OR)
\footnote{Restricted to true negatives (${n=565}$). 
${\mathrm{OR} = \mathrm{odds}(\mathrm{FP}\mid\mathrm{disagree})\;/\;\mathrm{odds}(\mathrm{FP}\mid\mathrm{agree})}$, with ${\mathrm{odds}(p)=p/(1-p)}$. ${\mathrm{OR} >1}$: more FPs on disputed cases.} range from 1.28 to 2.91 across models (per-model rates in Appendix~\ref{app:fpr}).

\begin{table}[h]
\centering
\begin{tabular}{lrrrr}
\toprule
\textbf{Subset} & \textbf{Cases} & \textbf{TN+FP} & \textbf{FP} & \textbf{FPR} \\
\midrule
Agree & 676 & 251 & 21 & 8.4\% \\
Disagree & 339 & 314 & 45 & 14.3\% \\
\midrule
Total & 1015 & 565 & 66 & 11.7\% \\
\bottomrule
\end{tabular}
\caption{False positives by annotator agreement. False positive rate FPR = FP / (TN+FP).}
\label{tab:fp_distribution}
\end{table}

Contrary to what one might expect, models do not become \textit{less} confident on difficult cases. They remain confident while being wrong. Confidence tracks accuracy quite well on agreed cases but not on disagreed ones (Table~\ref{tab:calibration_full}, Appendix~\ref{app:calibration}). Calibration Error is approximately twice as high on disagreement cases (0.30 versus 0.15 for the supervised ensemble). This pattern holds across all models. The concentration persists after controlling for chunk and article length and lexical overlap (logistic regression, Appendix~\ref{app:confounds}). Yet, we cannot definitively establish whether the residual difficulty reflects genuine legal ambiguity or other unmeasured factors. Nonetheless, the fact that 43\% of disagreements involve the fact versus law boundary (Table~\ref{tab:disagreement_structure}) is consistent with the hypothesis that some examples represent ambiguous applications of legal rules. A symmetric analysis of the false negatives shows the same direction but does not reach significance (Appendix~\ref{app:fn}), which is consistent with the very limited number of items tested (25).

\paragraph{Implications for Legal Practitioners}
Table~\ref{tab:ranking_practical} illustrates the implications of this ranking strategy from a practitioner's perspective. A legal professional researching how a specific Civil Code article is applied in practice would query the system and receive a ranked, top-$k$ list of court excerpts likely to apply it implicitly. With a cutoff set at a few hundred candidates, the unsupervised ensemble keeps precision high while roughly halving false positives compared to random sampling. Professionals could review a manageable shortlist rather than the full set. Moreover, the remaining false positives tend to be ambiguous cases. This pattern may be due to our adversarial filtering, which added difficult examples to the dataset. Even if it is the case, it suggests that professionals reviewing the flagged candidates would find borderline cases rather than obvious errors. Triage is necessary because the unfiltered test split alone contains 553,082 chunks. Manual review is unfeasible, whereas a ranked shortlist of a few hundred candidates is tractable for an expert.

\paragraph{Evaluation} Aggregate metrics can be deceptive when models behave differently depending on the level of difficulty. Splitting data based on annotator agreement makes it possible to assess whether improvements are limited to easy cases, where annotators agree, or also extend to more ambiguous cases.

\section{Qualitative Analysis}
\label{sec:qualitative}
We present three examples from the supervised ensemble illustrating when and why the model succeeds and fails (more examples in Appendix~\ref{app:examples}). Appendix~\ref{app:error_analysis} complements them with a quantitative error analysis of the supervised ensemble's 66 false positives. In 56\% of the cases, statutory language is present in the chunk but is not applied by the court. In 42\%, the model retrieves the right legal domain but an incorrect article.

\paragraph{True positive, agreement: Art.~1192 (contract interpretation)}
{\small A\textsubscript{1}=A\textsubscript{2}=Yes. Gold=Yes. Conf. 0.81 \checkmark}\\[2pt]
\textbf{Article} \textit{``On ne peut interpréter les clauses claires et précises à peine de dénaturation.''} [Clear and unambiguous contractual clauses may not be interpreted, as this would amount to distortion.]\\[1pt]
\textbf{Excerpt} \textit{``À l'inverse, sera écartée une contestation qui serait à l'évidence superficielle ou artificielle et le juge est tenu d'appliquer les clauses claires du contrat qui lui est soumis, si aucune interprétation n'en est nécessaire.''} [Conversely, a challenge that is clearly superficial or artificial will be dismissed, and the judge must apply the clear terms of the contract when no interpretation is required.]\\[1pt]
\textbf{Analysis} The court applies the principle of article~1192 without citing the number but both texts contain the same words (``clauses claires''). The logic according to which clear clauses require no interpretation is used and it is a clear implicit citation. Both annotators agreed and the model gave it a high confidence score.

\smallskip
\paragraph{False positive, disagreement: Art.~1361 (proof by writing)}
{\small A\textsubscript{1}=No, A\textsubscript{2}=Yes, A\textsubscript{3}=No (special regime). Gold=No. Conf.\ 0.87 $\times$}\\[2pt]
\textbf{Article} \textit{``Il peut être suppléé à l'écrit par l'aveu judiciaire, le serment décisoire ou un commencement de preuve par écrit corroboré par un autre moyen de preuve.''} [Written proof may be substituted by judicial admission, decisive oath, or a beginning of proof in writing corroborated by another means of proof.]\\[1pt]
\textbf{Excerpt} \textit{``L'offre de crédit produite n'est pas l'original du contrat, et la copie de cet acte juridique ne constitue ici qu'un commencement de preuve par écrit.''} [The credit offer produced is not the original contract, and the copy of this legal document constitutes here only a beginning of proof in writing.]\\[1pt]
\textbf{Analysis} A\textsubscript{2} noticed that the court invokes article 1361's concept of ``commencement de preuve par écrit'' to declare the evidentiary threshold not met. A\textsubscript{1} disagreed because consumer credit falls under the Code de la consommation, which has its own proof rules. A\textsubscript{3} sided with A\textsubscript{1}. In this case, the judge uses the same wording as the Civil Code but the applicable legal framework is different. The difficulty here is to determine where one regime ends and another begins.

\smallskip
\paragraph{True positive, disagreement: Art.~1219 (exception of non-performance)}
{\small A\textsubscript{1}=No, A\textsubscript{2}=Yes, A\textsubscript{3}=Yes. Gold=Yes. Conf.\ 0.71 \checkmark}\\[2pt]
\textbf{Article} \textit{``Une partie peut refuser d'exécuter son obligation, alors même que celle-ci est exigible, si l'autre n'exécute pas la sienne et si cette inexécution est suffisamment grave.''} [A party may refuse to perform its obligation, even if it is due, if the other party does not perform its own and if this non-performance is sufficiently serious.]\\[1pt]
\textbf{Excerpt} \textit{``Cependant, il n'est pas établi que la SARL NAJI AUTO est privée en permanence d'électricité depuis cette date et l'exception totale d'inexécution n'est donc pas justifiée.''} [However, it is not established that SARL NAJI AUTO has been permanently deprived of electricity since that date, and the total exception of non-performance is therefore not justified.]\\[1pt]
\textbf{Analysis} A\textsubscript{1}'s negative choice shows it is difficult to identify the proportionality test when the reasoning is mixed with factual finding. A\textsubscript{2} recognized that the judge applies the requirement that non-performance must be ``sufficiently serious'' to justify the suspension of one's obligations. Because the electricity deprivation is not established, the court concludes the exception is not justified. A\textsubscript{3} confirmed. The model detected the implicit citation despite the formulation.

\section{Conclusion}

Using a combination of computational methods and human annotation, we build a corpus for the detection of implicit citations in court decisions. Extensive experiments with state-of-the-art NLP methods highlight that the cases where legal experts disagree are also those where computational methods tend to fail. On the practical side, our ranking approach demonstrates that useful assistance may be possible even when perfect classification is not,
by surfacing candidates for human review rather than rendering binary judgments. One possibility consistent with our data but not proven by it, is that annotator disagreement reflects genuine ambiguity in the underlying task rather than  noise.
If so, the concentration of model errors on disputed cases may indicate a shared difficulty:
both humans and machines struggle with the same borderline instances.
This interpretation aligns with Hart's notion of legal ``open texture''  \citep{hart1961concept}. Whether this theoretical framework fully explains the observed pattern is left for future work.

\section*{Limitations}
\paragraph{Dataset size.} 1,015 examples is small by NLP standards but it is quite typical for expert-annotated legal datasets, and our statistical findings are solid. Annotation required advanced legal expertise in French law that cannot be crowd-sourced in order to adjudicate fine-grained distinctions between judicial reasoning and factual description.

\paragraph{Adversarial selection bias.} The only candidates retained were predicted positive by o3 because we used it as a filter. Reciprocally, our negatives are cases where o3 was wrong. Models that share reasoning patterns with it might be disadvantaged. Furthermore, models with different reasoning might perform better on our benchmark than on a randomly selected dataset. An annotation of 100 pairs that were rejected by o3 (\S \ref{par:bounding-selection-bias}) found no missed implicit applications, bounding this risk below 3\%. Moreover, the disagreement-failure pattern reproduces on architecturally distinct models (Appendices \ref{app:bge} and \ref{app:nli}), indicating it is not an artifact of o3 reasoning.

\paragraph{Zero-shot ensemble weights.} The weights in our unsupervised ensemble were set based on intuition rather than optimization. This preserves the unsupervised approach but performance could be improved with cross-validation tuning. A sensitivity analysis (Appendix \ref{app:ablation}) shows that the ranking is robust to weight changes.

\bibliography{custom}
\appendix

\section{Bi-Encoder Training Details}
\label{app:biencoder}

This appendix expands the retrieval pipeline of \S\ref{subsec:biencoder}. It details the sentence-aware chunking, explicit-pair extraction, data splits and FAISS indexing. 

\paragraph{Sentence-aware chunking} Motivation sections are segmented into chunks of up to 100 tokens using a custom algorithm. We split on sentence boundaries (\texttt{.}, \texttt{!}, \texttt{?}) while preserving legal abbreviations (e.g., \textit{C.~civ.}, \textit{art.}, \textit{al.}). Each chunk is made of up to two sentences if they fit within the token limit.

\paragraph{Explicit pair extraction} To identify chunks that explicitly cite Civil Code articles, we use complementary methods. We apply a set of regular expressions designed to capture common variations in French legal citations, such as article ranges (e.g., \emph{articles 1352 à 1352-9}), coordinated enumerations (\emph{et}), and abbreviated references to French law Codes (\emph{C.~civ.}, \emph{C.~com.}). We compute TF–IDF cosine similarity between each chunk and the corresponding article text. We discard candidate matches when their similarity falls below 0.15. It helps eliminate accidental lexical overlaps. For each retained positive pair, we construct a negative counterpart. We sample randomly an article that has a TF–IDF similarity with the chunk below 0.05. We keep these negatives separate from the training data. They are used exclusively for evaluation purposes.

\paragraph{Data splits} We split the 182,155 decisions into train (70\%), validation (15\%) and test (15\%) by \texttt{decision\_id}. We get 2,563,287 chunks for train, 555,658 for validation, and 553,082 for test. The bi-encoder is trained on 89,817 explicit positive pairs extracted from the training split (from 36,385 decisions). The validation set produces 19,642 positive pairs that were used for the selection of the threshold. The test split (553,082 chunks) is used exclusively for implicit candidate generation (\S\ref{sec:dataset}).

\paragraph{Model architecture} We use special tokens (\texttt{[ARTICLE]}, \texttt{[DECISION]}, \texttt{[LOI]}, \texttt{[MONTANT]}) to fine-tune JuriBERT.

\paragraph{FAISS indexing} We encode all 2,836 Civil Code articles (articles~1--2534). Embeddings are L2-normalized and indexed with \texttt{IndexFlatL2} for exact search. On validation pairs, 95\% of true matches fall within a squared L2 distance of $0.574$. This is the retrieval threshold we have chosen. 

\section{Article Renumbering}
\label{app:renumbering}
As mentioned in \S\ref{subsec:candidates}, the French Civil Code underwent a major reform in 2016, renumbering many articles in the contract law section (\emph{Ordonnance} n°2016-131 of February 10, 2016). For example, the former article 1134 (binding force of contracts) became article 1103. Court decisions from our corpus (2023--2025) may cite and use either the old or new numbering depending on when the underlying contract was formed. To avoid false implicit candidates, we constructed an equivalence table mapping pre-2016 article numbers to their post-2016 counterparts. When a chunk's candidate article is explicitly cited in the decision under either its old or new number, we exclude that candidate from the implicit pool. Our table covers approximately 150 article pairs and was compiled from official legislative sources. However, it is not exhaustive, as some articles were substantially modified rather than simply renumbered.

\section{Adversarial Filtering Prompt}
\label{app:o3prompt}
The adversarial filtering step of \S\ref{subsec:filtering} relied on the following prompt (in French, with English translation):

\begin{quote}\small
Tu es un juriste assistant spécialisé en droit français. Ta tâche consiste à déterminer si un extrait de décision judiciaire met en œuvre, applique ou reprend de manière implicite la règle de droit d'un article de loi donné, c'est-à-dire sans que le numéro ou la référence de l'article ne soit explicitement cité dans l'extrait, mais en reprenant son contenu, sa règle ou son principe.Réponds par 'OUI' ou 'NON' suivi d'une brève justification. IMPORTANT: Si tu n'es pas sûr ou qu'il y a un doute, réponds NON.\medskip

\textit{[You are a legal assistant with expertise in French law. Your task is to determine whether a court excerpt implicitly applies the rule contained in a given statutory article. The article must not be explicitly cited in the excerpt, but its substance, reasoning, or underlying principle may nonetheless be relied upon. Answer YES or NO, and provide a short justification. IMPORTANT: if there is any doubt, answer NO.]}\end{quote}

\section{Dataset Statistics}
\label{app:datasetstats}

The benchmark assembled in \S\ref{subsec:selection} is characterized here. The 1,015 pairs come from 829 decisions. An article appears on average 2.43 times (median 2), and 92\% appear five times or fewer.

Table \ref{tab:length_stats} reports the length statistics for chunks and articles. Chunks average 50 words in the benchmark (median 50, max 88). Articles have 60 words on average in the benchmark (median 46). The distribution of the benchmark is very similar to the one of the full Civil Code (mean 62, median 46).

\begin{table}[h]
\centering
\small
\setlength{\tabcolsep}{5pt}
\begin{tabular}{lrrrr}
\toprule
 & \textbf{Mean} & \textbf{Med.} & \textbf{Min} & \textbf{Max} \\
\midrule
Chunks (benchmark) & 49.9 & 50 & 14 & 88 \\
Articles (benchmark) & 60.2 & 46 & 7 & 564 \\
Articles (full Civil Code) & 61.5 & 46 & 3 & 883 \\
\bottomrule
\end{tabular}
\caption{Length statistics in words}
\label{tab:length_stats}
\end{table}

\section{Annotation Interface}
\label{app:interface}
The annotation study of \S\ref{sec:annotation} used the interface described here, designed to ease the task  while providing access to supplementary context when needed. For each entry, annotators were shown the candidate article together with its position in the Civil Code (book, title, chapter, and section), as well as the text chunk to be evaluated. Annotators could access the full court decision, in which the target chunk was visually highlighted. 

\paragraph{Annotator instructions.} The following instructions were given to the annotators:  \textit{L'article est-il appliqué implicitement ?} [\textit{Is the article implicitly applied?}]. 
The response options were: \textit{Oui} [Yes]; \textit{Non} [No]; \textit{Non, faits ou prétentions des parties uniquement} [No, facts or claims of the parties only]; \textit{Non, application d'un régime spécial} [No, application of a special regime]; \textit{Je ne sais pas} [I don't know]; \textit{À revoir} [To review].

\section{Structure of Disagreement}
\label{app:disagreement}

The disagreement structure summarized in \S\ref{sec:annotation} is detailed in Table~\ref{tab:disagreement_structure}, which reports how A\textsubscript{3} resolved each disputed case.

\begin{table}[h]
\centering
\begin{tabular}{lrr}
\toprule
\textbf{Resolution category (A\textsubscript{3})} & \textbf{N} & \textbf{\%} \\
\midrule
NO -- facts/party claims & 147 & 43.4\% \\
NO -- residual category & 106 & 31.3\% \\
NO -- special regime & 61 & 18.0\% \\
YES -- implicit application & 25 & 7.4\% \\
\bottomrule
\end{tabular}
\caption{Structure of disagreement cases resolved by Annotator A\textsubscript{3}.}
\label{tab:disagreement_structure}
\end{table}

\section{Grid Search Details}
\label{app:gridsearch}

The supervised encoders of \S\ref{sec:supervised} were tuned over the grid presented here:

\paragraph{Input configurations}
\begin{itemize}
    \item \texttt{cfg1}: \texttt{[article\_text] [SEP] [chunk]}
    \item \texttt{cfg2}: \texttt{[ARTICLE] [article\_text] [SEP] [CHUNK] [chunk]}
    \item \texttt{cfg3}: \texttt{[ARTICLE] Article \{num\}: [article\_text] [SEP] [CHUNK] [chunk]}
\end{itemize}

\paragraph{Embedding extraction} For encoders, we extract hidden states using four configurations: (i) mean pooling over last layer, (ii) mean pooling over penultimate layer (-2), (iii) mean pooling over average of last 4 layers (avg4), (iv) CLS token from last layer. For LLMs, we use configurations (i)--(iii) only.

\paragraph{Classification heads} Logistic regression (LR), and MLP with one or two hidden layers (MLP1, MLP2). MLP1 has one hidden layer of size 256. MLP2 has two hidden layers of sizes 256 and 64.

\paragraph{TF-IDF baseline} Unigrams and bigrams, with a minimum document frequency (\texttt{min\_df}) of 2, capped at \texttt{max\_features} of 50k, and sublinear TF scaling. IDF is computed on the training folds only. Logistic regression with SAGA solver, L2 regularization, and inverse regularization strength $C$ set to 1.0.

\paragraph{Best configurations} Table~\ref{tab:best-configs} reports the best configuration.
\begin{table}[h]
\centering
\small
\begin{tabular}{lcccc}
\toprule
\textbf{Encoder} & \textbf{Cfg} & \textbf{Pool} & \textbf{Layer} & \textbf{Head} \\
\midrule
SAUL-7B & cfg3 & mean & avg4 & LR \\
LLaMA-3.1-8B & cfg2 & mean & -2 & MLP1 \\
LawMA-8B & cfg1 & mean & last & MLP1 \\
CamemBERT & cfg3 & mean & avg4 & MLP2 \\
ST-MPNet & cfg1 & mean & avg4 & MLP2 \\
CamemBERTav2 & cfg1 & mean & avg4 & MLP1 \\
TF-IDF & cfg2 & -- & -- & LR \\
JuriBERT & cfg1 & cls & last & LR \\
ST-MiniLM & cfg3 & mean & -2 & MLP2 \\
\bottomrule
\end{tabular}
\caption{Best configuration per encoder (5-fold CV).}
\label{tab:best-configs}
\end{table}

\paragraph{Search space} We have a grid of 36 configurations per transformer and 27 per LLM. We use MCC calculated on out-of-fold predictions to select the best configuration.

\paragraph{Full metrics} Table~\ref{tab:supervised-full} presents the complete metrics for all supervised models.

\begin{table}[h]
\centering
\small
\begin{tabular}{lccccc}
\toprule
\textbf{Model} & \textbf{Thr} & \textbf{P} & \textbf{R} & \textbf{F1} & \textbf{MCC} \\
\midrule
SAUL-7B & .79 & .73 & .66 & .69 & .47 \\
LLaMA-3.1-8B & .61 & .73 & .65 & .69 & .46 \\
LawMA-8B & .84 & .79 & .54 & .64 & .46 \\
ST-MPNet & .63 & .76 & .54 & .63 & .43 \\
CamemBERT & .48 & .70 & .65 & .67 & .43 \\
CamemBERTav2 & .59 & .71 & .59 & .65 & .42 \\
TF-IDF & .47 & .73 & .56 & .63 & .41 \\
JuriBERT & .74 & .69 & .56 & .62 & .37 \\
ST-MiniLM & .54 & .69 & .55 & .61 & .37 \\
\midrule
Ensemble & .61 & .81 & .62 & .70 & .53 \\
\bottomrule
\end{tabular}
\caption{Supervised metrics (5-fold CV, positive class). Thr = threshold, P = precision, R = recall.}
\label{tab:supervised-full}
\end{table}

\paragraph{Confusion matrices} Table~\ref{tab:supervised-cm} reports confusion matrices.
\begin{table}[h]
\centering
\small
\begin{tabular}{lrrrr}
\toprule
\textbf{Model} & \textbf{TP} & \textbf{TN} & \textbf{FP} & \textbf{FN} \\
\midrule
Ensemble & 278 & 499 & \phantom{0}66 & 172 \\
SAUL-7B & 296 & 456 & 109 & 154 \\
LLaMA-3.1-8B & 291 & 457 & 108 & 159 \\
LawMA-8B & 241 & 501 & \phantom{0}64 & 209 \\
CamemBERT & 293 & 437 & 128 & 157 \\
CamemBERTav2 & 267 & 458 & 107 & 183 \\
TF-IDF & 251 & 473 & \phantom{0}92 & 199 \\
ST-MPNet & 244 & 486 & \phantom{0}79 & 206 \\
JuriBERT & 252 & 451 & 114 & 198 \\
ST-MiniLM & 248 & 454 & 111 & 202 \\
\bottomrule
\end{tabular}
\caption{Confusion matrices for supervised models (5-fold CV, optimized threshold).}
\label{tab:supervised-cm}
\end{table}

\section{Ensemble Search Details}
\label{app:ensemble}

We expand the ensemble search of \S\ref{sec:supervised}. We use nested cross-validations to avoid data leakage while evaluating ensemble methods. Folds are grouped by \texttt{decision\_id}. Different strategies are explored: logistic regression, weighted average and rank fusion. 

\paragraph{Top 5 configurations} Table~\ref{tab:ensemble-top5} reports the best ensembles.

\begin{table}[h]
\centering
\small
\setlength{\tabcolsep}{0.75pt} 
\begin{tabular}{llc}
\toprule
\textbf{Method} & \textbf{Models} & \textbf{MCC} \\
\midrule
Stacking-LR & Cam2+Juri+LLa+SAUL & .53 \\
Stacking-LR & Cam+Juri+SAUL+Mini+MP+Law & .53 \\
Stacking-LR & 
Cam+Juri+SAUL+Mini+MP+TF+Law & .53 \\
Weighted Avg & Cam2+Juri+SAUL+Law & .53 \\
Weighted Avg & Cam+Juri+SAUL+Mini+MP+Law & .53 \\
\bottomrule
\end{tabular}
\caption{Top 5 ensembles. Cam=CamemBERT, Cam2=CamemBERTav2, Juri=JuriBERT, LLa=LLaMA, Mini=ST-MiniLM, MP=ST-MPNet, TF=TF-IDF, Law=LawMA.}
\label{tab:ensemble-top5}
\end{table}

\paragraph{Ensemble metrics} Our retained ensemble stacks CamemBERTav2 + JuriBERT + LLaMA + SAUL with a logistic regression meta-learner. Table~\ref{tab:best-ensemble-metrics} reports detailed metrics at the optimized threshold.

\begin{table}[h]
\centering
\small
\begin{tabular}{lr}
\toprule
\textbf{Metric} & \textbf{Value} \\
\midrule
Models & Cam2+Juri+LLa+SAUL \\
Meta-learner & Logistic Regression \\
Threshold & .61 \\
\midrule
Accuracy & .77 \\
Balanced Accuracy & .75 \\
MCC & .53 \\
\midrule
Precision (yes/no) & .81 / .74 \\
Recall (yes/no) & .62 / .88 \\
F1 (yes/no) & .70 / .81 \\
\midrule
TP / TN / FP / FN & 278 / 499 / 66 / 172 \\
\bottomrule
\end{tabular}
\caption{Supervised ensemble metrics (5-fold nested CV).}
\label{tab:best-ensemble-metrics}
\end{table}

\section{Fine-tuned Cross-Encoder (BGE-reranker) Details}
\label{app:bge}

The fine-tuned BGE-reranker cross-encoder introduced in \S\ref{sec:supervised} is detailed here. We fine-tuned BGE-reranker-v2-m3 as a binary classifier on the chunk-article pairs under a 5-fold cross-validation for 3 epochs. We use AdamW and a learning rate of $2\times10^{-5}$ for the encoder body and $1\times10^{-4}$ for the classification head.

It is the strongest of the purpose-built baselines, with an accuracy of 0.72, an F1 of 0.69 and an MCC of 0.43. Its YES-rate (at 47.4\%) is close to the 44\% base rate. See Table~\ref{tab:bge}.

\begin{table}[h]
\centering
\small
\begin{tabular}{lc}
\toprule
\textbf{Metric} & \textbf{Value} \\
\midrule
Accuracy & 0.72 \\
F1 (positive) & 0.69 \\
MCC & 0.43 \\
YES-rate & 47.4\% \\
\midrule
OR (disputed versus consensual TN) & 2.22 \\
\quad 95\% CI & [1.51,\,3.28] \\
\quad $p$-value & 0.0001 \\
\bottomrule
\end{tabular}
\caption{Fine-tuned BGE-reranker-v2-m3}
\label{tab:bge}
\end{table}

\section{Entailment (NLI) Baseline Details} \label{app:nli} 

The entailment (NLI) baseline introduced in \S\ref{sec:supervised} is described here. We use XLM-RoBERTa-large fine-tuned on XNLI (\texttt{joeddav/xlm-roberta-large-xnli}) and we fine-tune it on our benchmark with the same grouped 5-fold cross-validation as the supervised classifiers in \S \ref{sec:supervised}. Each (chunk, article) pair is given to the model as a premise and a hypothesis. The premise is the court excerpt (chunk) and the hypothesis is the sentence \textit{``Ce passage applique la règle suivante : [article text]''} (``This passage applies the following rule: \ldots''). We tried three hypothesis templates and kept this one as the best performing, though the choice had little effect: all variants fell within 0.02 MCC of one another.

Fine-tuning uses AdamW at a learning rate of $1\times10^{-5}$, a batch size of 16, a maximum length of 512 tokens, for 4 epochs. Without fine-tuning, the model stays near chance (Table \ref{tab:nli}). On the other hand, the fine-tuning improves the MCC from 0.07 (zero-shot) to 0.35.

\begin{table}[h]
\centering
\setlength{\tabcolsep}{4pt}
\footnotesize
\begin{tabular}{lcccc}
\toprule
\textbf{Method} & \textbf{F1} & \textbf{Acc} & \textbf{MCC} & \textbf{OR} \\
\midrule
XNLI zero-shot  & .46 & .55 & .07 & ---  \\
XNLI fine-tuned & .63 & .68 & .35 & 2.26 \\
\midrule
TF-IDF          & .63 & .71 & .41 & 2.17  \\
Sup.\ ensemble  & .70 & .77 & .53 & 1.83  \\
\bottomrule
\end{tabular}
\caption{NLI baseline versus paper models (positive class).
TF-IDF and ensemble figures from Tables~\ref{tab:supervised}
and~\ref{tab:best-ensemble-metrics}.}
\label{tab:nli}
\end{table}

\section{Zero-Shot Prompts}
\label{app:prompt}

The zero-shot evaluation of \S\ref{sec:zeroshot} used the two prompts reproduced here:

\subsection{Standard Prompt}
This prompt was used for non-reasoning models:
\begin{quote}
\small
Tu es un expert en droit civil français. Ta tâche est de déterminer si un article du Code civil est implicitement appliqué dans un extrait de décision de justice.
IMPORTANT:
\begin{itemize}
    \item Réponds UNIQUEMENT par ``oui'' ou ``non''
    \item ``oui'' = l'article est implicitement appliqué (le raisonnement juridique utilise cet article sans le citer)
    \item ``non'' = l'article n'est pas appliqué (simple mention des faits, ou autre régime juridique)
\end{itemize}
Ne donne aucune explication, juste ``oui'' ou ``non''.
\medskip

\textit{[English: You are an expert in French civil law. Your task is to determine whether a Civil Code article is implicitly applied in a court decision excerpt. IMPORTANT: Answer ONLY ``yes'' or ``no''. ``yes'' = the article is implicitly applied (the legal reasoning uses this article without citing it). ``no'' = the article is not applied (mere mention of facts, or different legal regime). Provide no explanation, just ``yes'' or ``no''.]}
\end{quote}

\subsection{Reasoning Model Prompt}
For Qwen3-32B with thinking mode, we use a different prompt:
\begin{quote}
\small
Tu es un expert en droit civil français. Ta tâche est de déterminer si un article du Code civil est implicitement appliqué dans un extrait de décision de justice.

Analyse l'extrait en considérant:
\begin{enumerate}
    \item Le raisonnement juridique utilisé par le juge
    \item Les concepts juridiques mobilisés (même sans citation explicite)
    \item La cohérence entre l'article proposé et le raisonnement de la décision
\end{enumerate}

À la fin de ton analyse, conclus OBLIGATOIREMENT par une ligne contenant uniquement:\\
RÉPONSE: oui\\
ou\\
RÉPONSE: non

\begin{itemize}
    \item ``oui'' = l'article est implicitement appliqué (le raisonnement juridique utilise cet article sans le citer)
    \item ``non'' = l'article n'est pas appliqué (simple mention des faits, ou autre régime juridique par exemple)
\end{itemize}
\medskip

\textit{[English: You are an expert in French civil law. Your task is to figure out if a Civil Code article is implicitly applied in a court decision excerpt. Analyze the excerpt by looking at: (1) The legal reasoning the judge uses, (2) The legal concepts involved (even if not explicitly cited), (3) Whether the proposed article actually fits with what the decision is saying. At the end of your analysis, you MUST conclude with a line containing only: ANSWER: yes, or, ANSWER: no. ``yes'' = the article is implicitly applied (the legal reasoning relies on this article without citing it). ``no'' = the article is not applied (just factual mention, or different legal framework for instance).]}
\end{quote}

\section{Zero-Shot Full Results}
\label{app:zeroshot}

These tables give the complete zero-shot results behind \S\ref{sec:zeroshot}: Table~\ref{tab:zeroshot-gold} presents the zero-shot metrics against the gold standard while  Tables~\ref{tab:zeroshot-a1} and \ref{tab:zeroshot-a2} showcase the individual annotators.

\begin{table}[h]
\centering
\fontsize{10pt}{12pt}\selectfont
\setlength{\tabcolsep}{3pt}
\begin{tabular}{lccccccc}
\toprule
\textbf{Model} & \textbf{Yes} & \textbf{P} & \textbf{R} & \textbf{F1} & \textbf{Acc} & \textbf{MCC} \\
\midrule
Qwen2.5-32B & 66 & 55 & 82 & 66 & 63 & 31 \\
Qwen2.5-7B & 54 & 57 & 70 & 63 & 63 & 28 \\
LLaMA-3.1-8B & 22 & 70 & 35 & 47 & 65 & 28 \\
LLaMA-3.1-70B$^\dagger$ & 84 & 50 & 94 & 65 & 55 & 25 \\
Mistral-Nemo-12B & 75 & 50 & 85 & 63 & 56 & 21 \\
Gemma-2-27B & 94 & 46 & 98 & 63 & 49 & 17 \\
Qwen3-32B$^\ddagger$ & 85 & 48 & 91 & 63 & 52 & 16 \\
Aya-Expanse-32B & 94 & 45 & 96 & 62 & 47 & \phantom{0}8 \\
Command-R-35B & 96 & 45 & 97 & 61 & 46 & \phantom{0}4 \\
SAUL-7B & 95 & 45 & 95 & 61 & 46 & \phantom{0}3 \\
\bottomrule
\end{tabular}
\caption{Zero-shot results vs.\ gold ($n=1{,}015$), positive class, in \%. $^\dagger$4-bit quantization. $^\ddagger$Reasoning model.}
\label{tab:zeroshot-gold}
\end{table}

\begin{table}[h]
\centering
\fontsize{10pt}{12pt}\selectfont
\setlength{\tabcolsep}{4pt}
\begin{tabular}{lccccccc}
\toprule
\textbf{Model} & \textbf{Yes} & \textbf{P} & \textbf{R} & \textbf{F1} & \textbf{Acc} & \textbf{MCC} \\
\midrule
Qwen2.5-32B & 66 & 61 & 80 & 69 & 64 & 30 \\
Qwen2.5-7B & 54 & 62 & 67 & 64 & 63 & 26 \\
LLaMA-3.1-70B$^\dagger$ & 84 & 55 & 93 & 69 & 59 & 24 \\
LLaMA-3.1-8B & 22 & 71 & 31 & 44 & 59 & 23 \\
Mistral-Nemo-12B & 75 & 55 & 83 & 66 & 58 & 18 \\
Qwen3-32B$^\ddagger$ & 85 & 54 & 91 & 67 & 56 & 17 \\
Gemma-2-27B & 94 & 52 & 97 & 68 & 54 & 15 \\
Aya-Expanse-32B & 94 & 51 & 96 & 66 & 52 & \phantom{0}7 \\
Command-R-35B & 96 & 50 & 97 & 66 & 51 & \phantom{0}4 \\
SAUL-7B & 95 & 51 & 95 & 66 & 51 & \phantom{0}4 \\
\bottomrule
\end{tabular}
\caption{Zero-shot results vs.\ A\textsubscript{1}, positive class, in \%. $^\dagger$4-bit quantization. $^\ddagger$Reasoning model.}
\label{tab:zeroshot-a1}
\end{table}

\begin{table}[h]
\centering
\fontsize{10pt}{12pt}\selectfont
\setlength{\tabcolsep}{4pt}
\begin{tabular}{lccccccc}
\toprule
\textbf{Model} & \textbf{Yes} & \textbf{P} & \textbf{R} & \textbf{F1} & \textbf{Acc} & \textbf{MCC} \\
\midrule
Qwen2.5-32B & 66 & 76 & 74 & 75 & 67 & 26 \\
Qwen2.5-7B & 54 & 75 & 60 & 67 & 60 & 18 \\
LLaMA-3.1-70B$^\dagger$ & 84 & 71 & 88 & 79 & 68 & 17 \\
Gemma-2-27B & 94 & 69 & 97 & 81 & 69 & 17 \\
Mistral-Nemo-12B & 75 & 71 & 80 & 76 & 65 & 16 \\
LLaMA-3.1-8B & 22 & 81 & 27 & 40 & 47 & 16 \\
Qwen3-32B$^\ddagger$ & 85 & 70 & 88 & 78 & 67 & 14 \\
SAUL-7B & 95 & 68 & 96 & 79 & 67 & \phantom{0}6 \\
Command-R-35B & 96 & 67 & 97 & 79 & 67 & \phantom{0}4 \\
Aya-Expanse-32B & 94 & 67 & 94 & 78 & 65 & \phantom{0}1 \\
\bottomrule
\end{tabular}
\caption{Zero-shot results vs.\ A\textsubscript{2}, positive class, in \%. $^\dagger$4-bit quantization. $^\ddagger$Reasoning model.}
\label{tab:zeroshot-a2}
\end{table}

\section{Retrieval-Augmented Few-Shot Prompting}
\label{app:fewshot}

Few-shot prompting, mentioned in \S\ref{sec:zeroshot}, is evaluated on the three strongest zero-shot models: Qwen-2.5-32B, Qwen-2.5-7B and LLaMA-3.1-8B. Table~\ref{tab:fewshot} shows that few-shot prompting never approaches the supervised ensemble (MCC = 0.53).

For each query we retrieve $k/2$ positive and $k/2$ negative in-context examples by BM25 from the training fold (5-fold CV). We exclude the query's article. Few-shot brings at most +0.02 MCC for Qwen-2.5-7B. It degrades LLaMA-3.1-8B by pushing its YES-rate from 22\% toward 50\% without improving discrimination.

\begin{table}[h]
\centering
\begin{tabular}{lcccc}
\toprule
\textbf{Model} & \textbf{$k$=0} & \textbf{$k$=2} & \textbf{$k$=4} & \textbf{$k$=8} \\
\midrule
Qwen-2.5-32B  & .309 & .306 & .324 & .324 \\
Qwen-2.5-7B   & .275 & .294 & .267 & .248 \\
LLaMA-3.1-8B  & .284 & .213 & .210 & .212 \\
\bottomrule
\end{tabular}
\caption{Few-shot MCC on gold ($n=1{,}015$; Qwen-2.5-32B at $k{=}8$: $n=1{,}008$). $k$ in-context examples retrieved by BM25.}
\label{tab:fewshot}
\end{table}

\section{Unsupervised Ensemble: Weights and Sensitivity}
\label{app:ablation}

We present the heuristic of \S\ref{sec:unsupervised} and report the sensitivity study. The ranking score is:
\[
\begin{aligned}
S ={}& 0.25\sum_{i=1}^{4} v_i
+ \sum_{j=2}^{4}\beta_j\,
\mathbf{1}\!\left[\sum_i v_i \ge j\right] \\
&+ 0.2\,\mathrm{TF\text{-}IDF}
+ 0.2\,\mathrm{BM25}
+ 0.4\,\mathrm{CE}.
\end{aligned}
\]
where ${v_i \in \{0,1\}}$ is LLM $i$'s vote. ${\beta_2=0.3}$, ${\beta_3=0.5}$ and ${\beta_4=1.0}$ are cumulative so that the unanimous bonus is ${0.3+0.5+1.0=1.8}$. TF-IDF, BM25 and cross-encoder (CE) scores are normalized to ${[0,1]}$. Ties are broken by ${\varepsilon\cdot\text{CE}}$ with ${\varepsilon = 10^{-6}}$.

We test whether the ranking score is sensitive to the exact choice of weights. Overall, it is not, as the following results show:
\begin{itemize}
    \item Scaling the LLM block by a factor $\lambda$ keeps AP almost unchanged across a wide range of values. AP is ${0.668 \pm 0.002}$ for ${\lambda \in [0.25,10]}$ (P@200 = 0.755). However, entirely removing the LLM block does affect performance, since it then drops (${\lambda=0}$, AP = 0.604). ${\lambda\to\infty}$ is the symmetric ablation. When we keep the LLM block only (we set the TF-IDF, BM25 and cross-encoder weights to zero), we notice a slight increase ($\lambda \to \infty$, AP = 0.675).
    \item Perturbing all weights randomly by factors in ${[0.5,1.5]}$ over 200 configurations shows an AP standard deviation of only 0.001.
    \item Removing any single component changes AP by at most 0.005.
\end{itemize}

\section{Unsupervised Ranking Details}
\label{app:ranking}

Complementing the unsupervised ranking of \S\ref{sec:unsupervised}, this appendix gives the tie-breaking rule and the average-precision scores for each annotator.

\paragraph{Tie-breaking} There were ties because of the binary outputs (0 or 1) of the LLMs. To break them we used a trace from the cross-encoder. The final score is $\text{Score}_{\text{final}} = \text{Score}_{\text{LLMs}} + \varepsilon \times \text{Score}_{\text{CE}}$, with $\varepsilon = 10^{-6}$. 

\paragraph{Full results by annotator} 
Figure~\ref{fig:gain_vs_random} (main text) reports the gain in true positives on gold labels. Figures~\ref{fig:gain_vs_random_a1} and \ref{fig:gain_vs_random_a2} show the same curves for A\textsubscript{1} and A\textsubscript{2} individually.

Table~\ref{tab:ranking_by_annotator} reports AP for each annotator separately. A\textsubscript{2} says yes more often (67\% positive) and thus shows higher AP values. The relative ranking of methods is nonetheless consistent across annotators.

\begin{table}[h]
\centering
\small
\begin{tabular}{lccc}
\toprule
\textbf{Method} & \textbf{A\textsubscript{1}}  & \textbf{A\textsubscript{2}} & \textbf{Gold} \\
\midrule
Random & .50 & .67 & .44 \\
TF-IDF & .61 & .74 & .57 \\
BM25 & .61 & .73 & .57 \\
Cross-Encoder & .63 & .76 & .61 \\
\midrule
LLM\_Inter4 & .67 & .78 & .66 \\
Unsupervised Ensemble & .69 & .79 & .67 \\
\bottomrule
\end{tabular}
\caption{Average Precision by annotator for all ranking methods.}
\label{tab:ranking_by_annotator}
\end{table}

\paragraph{Coverage analysis} Tables~\ref{tab:ranking_recall_pct} and \ref{tab:ranking_recall_raw} display recall at diverse cutoffs. At $k=300$ (29.6\% of the corpus), the unsupervised ensemble retrieves 47\% of all true positives. Table~\ref{tab:ranking_fp_main} details the FP/disagreement breakdown.

\begin{table}[h]
\centering
\small
\setlength{\tabcolsep}{4pt}
\begin{tabular}{lcccc}
\toprule
$k$ & \textbf{Unsup. Ensemble} & \textbf{Inter4} & \textbf{CrossEnc.} & \textbf{Random} \\
\midrule
100 & 17\% & 17\% & 16\% & 10\% \\
200 & 34\% & 33\% & 30\% & 20\% \\
300 & 47\% & 47\% & 42\% & 30\% \\
500 & 69\% & 65\% & 62\% & 49\% \\
\bottomrule
\end{tabular}
\caption{Recall at $k$ (\%) for gold labels.}
\label{tab:ranking_recall_pct}
\end{table}

\begin{table}[h]
\centering
\small
\setlength{\tabcolsep}{4pt}
\begin{tabular}{lcccc}
\toprule
$k$ & \textbf{Unsup. Ensemble} & \textbf{Inter4} & \textbf{CrossEnc.} & \textbf{Random} \\
\midrule
100 & \phantom{0}77 & \phantom{0}76 & \phantom{0}72 & \phantom{0}44 \\
200 & 151 & 150 & 137 & \phantom{0}89 \\
300 & 211 & 213 & 189 & 133 \\
500 & 310 & 294 & 281 & 222 \\
\bottomrule
\end{tabular}
\caption{True positives retrieved at $k$ (out of 450 total).}
\label{tab:ranking_recall_raw}
\end{table}

\paragraph{Results by annotator} Figures~\ref{fig:gain_vs_random_a1} and \ref{fig:gain_vs_random_a2} show the gain in true positives for A\textsubscript{1} and A\textsubscript{2}. The unsupervised ensemble is the best method for both annotators. The gain in absolute value changes with the annotator.

\begin{figure}[h]
\centering
\begin{tikzpicture}
\begin{axis}[
    width=0.85\columnwidth,
    height=5.5cm,
    xlabel={$k$},
    ylabel={Gain vs Random (TP)},
    xmin=0, xmax=1050,
    ymin=-5, ymax=85,
    xtick={0,200,400,600,800,1000},
    xticklabel style={font=\scriptsize},
    legend style={
        font=\scriptsize,
        at={(0.5,-0.35)},
        anchor=north,
        legend columns=2,
        cells={anchor=west},
    },
    grid=major,
    grid style={dashed,gray!30},
    title={A\textsubscript{1}},
]
\addplot[blue, mark=*, thick, mark size=1.5pt] coordinates {
    (50,10) (100,18) (200,31) (300,43) (400,50) (500,44) (600,37) (800,27) (1000,2)
};
\addlegendentry{TF-IDF}
\addplot[green!60!black, mark=*, thick, mark size=1.5pt] coordinates {
    (50,10) (100,23) (200,41) (300,47) (400,45) (500,51) (600,44) (800,23) (1000,6)
};
\addlegendentry{CrossEncoder}
\addplot[magenta, mark=*, thick, mark size=1.5pt] coordinates {
    (50,16) (100,30) (200,51) (300,64) (400,73) (500,79) (600,71) (800,50) (1000,7)
};
\addlegendentry{Unsup. Ensemble}
\addplot[brown, mark=*, thick, mark size=1.5pt] coordinates {
    (50,16) (100,29) (200,49) (300,65) (400,62) (500,61) (600,56) (800,29) (1000,6)
};
\addlegendentry{LLM\_Inter4}
\end{axis}
\end{tikzpicture}
\caption{Gain in true positives vs.\ random ranking (A\textsubscript{1}).}
\label{fig:gain_vs_random_a1}
\end{figure}

\begin{figure}[h]
\centering
\begin{tikzpicture}
\begin{axis}[
    width=0.85\columnwidth,
    height=5.5cm,
    xlabel={$k$},
    ylabel={Gain vs Random (TP)},
    xmin=0, xmax=1050,
    ymin=-5, ymax=65,
    xtick={0,200,400,600,800,1000},
    xticklabel style={font=\scriptsize},
    legend style={
        font=\scriptsize,
        at={(0.5,-0.35)},
        anchor=north,
        legend columns=2,
        cells={anchor=west},
    },
    grid=major,
    grid style={dashed,gray!30},
    title={A\textsubscript{2}},
]
\addplot[blue, mark=*, thick, mark size=1.5pt] coordinates {
    (50,6) (100,12) (200,23) (300,35) (400,36) (500,34) (600,25) (800,18) (1000,1)
};
\addlegendentry{TF-IDF}
\addplot[green!60!black, mark=*, thick, mark size=1.5pt] coordinates {
    (50,9) (100,18) (200,32) (300,32) (400,33) (500,29) (600,37) (800,25) (1000,5)
};
\addlegendentry{CrossEncoder}
\addplot[magenta, mark=*, thick, mark size=1.5pt] coordinates {
    (50,11) (100,19) (200,36) (300,49) (400,53) (500,56) (600,54) (800,39) (1000,8)
};
\addlegendentry{Unsup. Ensemble}
\addplot[brown, mark=*, thick, mark size=1.5pt] coordinates {
    (50,10) (100,19) (200,36) (300,47) (400,44) (500,36) (600,41) (800,29) (1000,5)
};
\addlegendentry{LLM\_Inter4}
\end{axis}
\end{tikzpicture}
\caption{Gain in true positives vs.\ random ranking (A\textsubscript{2}).}
\label{fig:gain_vs_random_a2}
\end{figure}

\begin{table}[h]
\centering
\small
\begin{tabular}{lccc}
\toprule
$k$ & \textbf{FP (Agree)} & \textbf{FP (Disagree)} & \textbf{\% Disagree} \\
\midrule
50 & \phantom{0}2 & 10 & 83\% \\
100 & \phantom{0}8 & 15 & 65\% \\
200 & 20 & 29 & 59\% \\
\bottomrule
\end{tabular}
\caption{False positive origin by annotator agreement. Disagreement cases (33\% of data) account for 59--83\% of FPs until $k$ = 200.}
\label{tab:ranking_fp_main}
\vspace{-\baselineskip}
\end{table}

\section{Per-Model False Positive Rates}
\label{app:fpr}
These are the per-model false-positive rates behind the odds ratios reported in \S\ref{sec:fp_analysis}. Table~\ref{tab:fpr_analysis} reports false positive rates on agreement versus disagreement  for all individual models and the supervised ensemble. Figure~\ref{fig:fpr_by_agreement} (main text) represents the same data.

\begin{table}[h]
\centering
\small
\begin{tabular}{lcccc}
\toprule
\textbf{Model} & \textbf{FPR\textsubscript{ag}} & \textbf{FPR\textsubscript{dis}} & \textbf{OR} & \textbf{$p$} \\
\midrule
LawMA-8B & \phantom{0}6.0\% & 15.6\% & 2.91 & $<$.001 \\
ST-MPNet & \phantom{0}8.0\% & 18.8\% & 2.67 & $<$.001 \\
CamemBERT & 14.3\% & 29.3\% & 2.48 & $<$.001 \\
LLaMA-3.1-8B & 12.7\% & 24.2\% & 2.19 & $<$.001 \\
TF-IDF & 10.8\% & 20.7\% & 2.17 & .001 \\
JuriBERT & 14.3\% & 24.8\% & 1.97 & .002 \\
SAUL-7B & 14.7\% & 22.9\% & 1.72 & .013 \\
CamemBERTav2 & 15.5\% & 21.7\% & 1.50 & .058 \\
ST-MiniLM & 17.5\% & 21.3\% & 1.28 & .253 \\
\midrule
Ensemble & \phantom{0}8.4\% & 14.3\% & 1.83 & .030 \\
\bottomrule
\end{tabular}
\caption{False positive rates on agreement (FPR\textsubscript{ag}, $n=251$) vs.\ disagreement (FPR\textsubscript{dis}, $n=314$) subsets. Seven models and the ensemble significant after FDR correction ($p < 0.05$).}
\label{tab:fpr_analysis}
\vspace{-\baselineskip}
\end{table}

\section{Calibration Details}
\label{app:calibration}

The calibration gap discussed in \S\ref{sec:discussion} is detailed in Table~\ref{tab:ece_summary}, which reports Expected Calibration Error (ECE) for all models. It is computed separately on agreement and disagreement cases. All models show higher ECE on disagreement cases. Table~\ref{tab:calibration_full} provides a breakdown by confidence bin for the retained supervised ensemble.

\begin{table}[h]
\centering
\small
\begin{tabular}{lccc}
\toprule
\textbf{Model} & \textbf{ECE\textsubscript{ag}} & \textbf{ECE\textsubscript{dis}} & \textbf{$\Delta$} \\
\midrule
TF-IDF & .17 & .34 & + .17 \\
ST-MiniLM & .16 & .31 & + .15 \\
Ensemble & .15 & .30 & + .15 \\
CamemBERTav2 & .15 & .30 & + .14 \\
ST-MPNet & .16 & .30 & + .14 \\
CamemBERT & .16 & .29 & + .13 \\
LLaMA-3.1-8B & .17 & .28 & + .11 \\
JuriBERT & .21 & .32 & + .11 \\
Lawma-8B & .18 & .28 & + .10 \\
SAUL-7B & .20 & .29 & + .09 \\
\bottomrule
\end{tabular}
\caption{Expected Calibration Error by agreement status.}
\label{tab:ece_summary}
\end{table}

\begin{table}[h]
\centering
\small
\begin{tabular}{l|ccc|ccc}
\toprule
 & \multicolumn{3}{c|}{\textbf{Agreement}} & \multicolumn{3}{c}{\textbf{Disagreement}} \\
\textbf{Bin} & {Acc.} & Gap & n & {Acc.} & Gap & n \\
\midrule
0.1--0.2 & .26 & + .12 & 175 & .03 & $-$.12 & 117 \\
0.2--0.3 & .39 & + .14 & 66 & .06 & $-$ .19 & \phantom{0}54 \\
0.3--0.4 & .54 & + .19 & 65 & .00 & $-$ .35 & \phantom{0}33 \\
0.4--0.5 & .70 & + .25 & 43 & .07 & $-$ .38 & \phantom{0}43 \\
0.5--0.6 & .55 & + .01 & 38 & .07 & $-$ .48 & \phantom{0}29 \\
0.6--0.7 & .82 & + .16 & 73 & .13 & $-$ .51 & \phantom{0}15 \\
0.7--0.8 & .94 & + .18 & 93 & .12 & $-$ .64 & \phantom{0}24 \\
0.8--0.9 & .98 & + .13 & 122 & .38 & $-$ .47 & \phantom{0}24 \\
\bottomrule
\end{tabular}
\caption{Full calibration for supervised ensemble.}
\label{tab:calibration_full}
\end{table}

\section{Controlling for Surface Confounds}
\label{app:confounds}

To support the claim in \S\ref{sec:discussion} that error concentration reflects a genuine effect rather than a surface artifact, we test whether disagreement-related errors can instead be explained by surface features. We restrict our analysis to true negatives (gold~=~NO, ${n=565}$). We regress a false-positive indicator on a disagreement indicator. We test three nested specifications: \textsc{M0} (disagreement only), \textsc{M1} (+~log chunk and article length), and \textsc{M2} (+~TF-IDF overlap). The groups have very similar surface statistics. The disagreement odds ratio barely moves once the controls are added (Table~\ref{tab:confounds}): at most 0.08 across models, and 1.83~$\to$~1.85 for the ensemble. Surface properties do not explain the concentration of errors on disputed cases.

\begin{table}[h]
\centering
\small
\begin{tabular}{lccc}
\toprule
\textbf{Model} & \textbf{M0} & \textbf{M1} & \textbf{M2} \\
\midrule
LawMA-8B      & 2.91 & 2.91 & 2.92 \\
ST-MPNet      & 2.67 & 2.70 & 2.75 \\
CamemBERT     & 2.47 & 2.52 & 2.53 \\
LLaMA-3.1-8B  & 2.19 & 2.19 & 2.21 \\
TF-IDF        & 2.17 & 2.16 & 2.17 \\
JuriBERT      & 1.97 & 1.97 & 1.97 \\
SAUL-7B       & 1.72 & 1.72 & 1.72 \\
CamemBERTav2  & 1.50 & 1.50 & 1.51 \\
ST-MiniLM     & 1.28 & 1.27 & 1.27 \\
\midrule
Ensemble      & 1.83 & 1.83 & 1.85 \\
\bottomrule
\end{tabular}
\caption{Odds ratio of false positive for disagreement versus agreement on true negatives (${n=565}$)}
\label{tab:confounds}
\end{table}

\section{False Negatives by Agreement}
\label{app:fn}

The symmetric false-negative analysis noted in \S\ref{sec:discussion} is reported in Table~\ref{tab:fn_distribution}. We repeat the agreement/disagreement split on false negatives (gold~=~YES). The supervised ensemble has FNR 37.9\% on agreement versus 44.0\% on disagreement (OR~=~1.29, ${p = 0.53}$). The direction matches the false-positive analysis. However, there are only 25 positives within the disagreement subset. This is a structural consequence of the 92.6\% of disputed cases resolving to NO. As a consequence, statistical power is limited. Six of nine individual classifiers show ${\mathrm{OR} >1}$. None is individually significant.

\begin{table}[h]
\centering
\small
\begin{tabular}{lrrrr}
\toprule
\textbf{Subset} & \textbf{Cases} & \textbf{TP+FN} & \textbf{FN} & \textbf{FNR} \\
\midrule
Agree & 676 & 425 & 161 & 37.9\% \\
Disagree & 339 & \phantom{0}25 & \phantom{0}11 & 44.0\% \\
\bottomrule
\end{tabular}
\caption{False negatives by annotator agreement (gold = YES). FNR = FN / (TP+FN).}
\label{tab:fn_distribution}
\end{table}

\section{Additional Qualitative Examples}
\label{app:examples}

Beyond the three cases analyzed in \S\ref{sec:qualitative}, this appendix presents six annotated examples. All predictions were made using the supervised ensemble (Stacking-LR: CamemBERTav2 + JuriBERT + LLaMA + SAUL).

\subsection{True positive, agreement: Art.~2274 (good faith presumption)}
{\small A\textsubscript{1}=A\textsubscript{2}=Yes. Gold=Yes. Conf.\ 0.75 \checkmark}\\[2pt]
\textbf{Article} \textit{``La bonne foi est toujours présumée, et c'est à celui qui allègue la mauvaise foi à la prouver.''} [Good faith is always presumed, and it is for the party alleging bad faith to prove it.]\\[1pt]
\textbf{Excerpt} \textit{``Néanmoins, ce seul constat de l'accroissement de la dette locative ne suffit pas à caractériser par lui seul la mauvaise foi de la débitrice.''} [Nevertheless, the mere observation of the increase in rental debt is not sufficient in itself to establish the bad faith of the debtor.]\\[1pt]
\textbf{Analysis} The presumption of the article 2274 is applied by the court that requires affirmative proof of bad faith. Circumstantial evidence (merely observing increased debt) does not qualify. Both annotators recognized the application and the model identified it properly with satisfying confidence.

\smallskip
\subsection{True positive, agreement: Art.~1310 (solidarity not presumed)}
{\small A\textsubscript{1}=A\textsubscript{2}=Yes. Gold=Yes. Conf.\ 0.71 \checkmark}\\[2pt]
\textbf{Article} \textit{``La solidarité est légale ou conventionnelle; elle ne se présume pas.''} [Solidarity is statutory or contractual; it may not be presumed.]\\[1pt]
\textbf{Excerpt} \textit{``La solidarité ne se présume pas et qu'aucun texte ne l'institue pour le paiement des charges de copropriété du seul fait de la nature de cette dette.''} [Solidarity is not presumed, and no text establishes it for the payment of co-ownership charges solely by virtue of the nature of this debt.]\\[1pt]
\textbf{Analysis} The same words as the article 1310 (``ne se présume pas'') are used by the court. It rejects the existence of solidarity between co-owners and also outlines that no statutory provision establishes solidarity. Doing so, it reaffirms that solidarity requires an explicit legal or contractual basis. Both annotators agreed on an implicit citation. The model was right as well.

\smallskip
\subsection{True negative, agreement: Art.~1241 (tort liability)}
{\small A\textsubscript{1}=A\textsubscript{2}=No. Gold=No. Conf.\ 0.27 \checkmark}\\[2pt]
\textbf{Article} \textit{``Chacun est responsable du dommage qu'il a causé non seulement par son fait, mais encore par sa négligence ou par son imprudence.''} [Every person is liable for the damage they have caused not only by their own act, but also by their negligence or imprudence.]\\[1pt]
\textbf{Excerpt} \textit{``Le principe du droit à indemnisation intégrale du demandeur est en revanche contesté par l'assureur qui reproche à M. [F] [S] d'avoir commis des fautes à l'origine de son dommage.''} [The principle of the plaintiff's right to full compensation is, however, contested by the insurer, who accuses Mr. [F] [S] of having committed faults that caused his damage.]\\[1pt]
\textbf{Analysis} There are some semantic similarity (``fautes,'' ``dommage''), but this extract reports the insurer's position and not a judicial reasoning. The court only describes what a party claims. Both annotators identified this as a factual chunk and the model rightfully assigned low confidence.

\smallskip
\subsection{False positive, disagreement: Art.~2251 (renunciation of prescription)}
{\small A\textsubscript{1}=Yes, A\textsubscript{2}=No, A\textsubscript{3}=No. Gold=No. Conf.\ 0.69 $\times$}\\[2pt]
\textbf{Article} \textit{``La renonciation à la prescription est expresse ou tacite. La renonciation tacite résulte de circonstances établissant sans équivoque la volonté de ne pas se prévaloir de la prescription.''} [Prescription may be renounced expressly or tacitly. A tacit renunciation is inferred from circumstances that unequivocally show an intent to forgo the defense.]\\[1pt]
\textbf{Excerpt} \textit{``Elle conteste toute renonciation à son droit de se prévaloir de la forclusion de l'action des époux [R] qui ne peut se présumer et doit résulter, lorsqu'elle est tacite, d'une volonté de renoncer non équivoque.''} [She denies having waived her right to raise the time-bar against the [R] spouses' claim. This waiver cannot be presumed; if tacit, it must stem from an unequivocal intent to give up that defense.]\\[1pt]
\textbf{Analysis} A\textsubscript{1} noticed the legal standard being invoked but A\textsubscript{2} and A\textsubscript{3} noted that ``Elle conteste...'' is a party's argument. The court reports what one party argues, not applying the rule itself. The model incorrectly predicted this case. Cases like this illustrate that it can be difficult to distinguish the description of facts and party claims from the actual application of law to those facts.

\smallskip
\subsection{False negative, disagreement: Art.~1118 (acceptance of offer)}
{\small A\textsubscript{1}=Yes, A\textsubscript{2}=No, A\textsubscript{3}=Yes. Gold=Yes. Conf.\ 0.47 $\times$}\\[2pt]
\textbf{Article} \textit{``L'acceptation est la manifestation de volonté de son auteur d'être lié dans les termes de l'offre...''} [Acceptance is the expression of its author's intention to be bound by the terms of the offer...]\\[1pt]
\textbf{Excerpt} \textit{``Cette affirmation est exacte, mais la bonne réception de la demande de Monsieur [E] [L] n'en signifie pas pour autant l'acceptation.''} [This assertion is correct, but the proper receipt of Mr. [E] [L]'s request does not thereby signify acceptance.]\\[1pt]
\textbf{Analysis} Article 1118 core principle is the distinction between receipt and acceptance. It is a key concept of contract formation. Receiving a communication is different from acceptance without a manifestation of intent to be bound. A\textsubscript{2} may have focused on the absence of the article characteristic vocabulary. A\textsubscript{1} and A\textsubscript{3} recognized the underlying legal distinction being applied. The model was wrong but exhibits moderate confidence. Detecting implicit citations is more difficult when the statutory language is not echoed in the decision.

\smallskip
\subsection{False positive, agreement: Art.~606 (major repairs)}
{\small A\textsubscript{1}=A\textsubscript{2}=No. Gold=No. Conf.\ 0.80 $\times$}\\[2pt]
\textbf{Article} \textit{``Les grosses réparations sont celles des gros murs et des voûtes, le rétablissement des poutres et des couvertures entières...''} [Major repairs comprise those to main walls and vaults, the restoration of beams and entire roofs...]\\[1pt]
\textbf{Excerpt} \textit{``Le bail commercial précise au titre de l'entretien et des réparations que « le bailleur aura à sa charge les réparations afférentes aux gros murs et voûtes, le rétablissement des poutres et des couvertures entières...''} [The commercial lease specifies, under the heading of maintenance and repairs, that the lessor shall bear the cost of repairs relating to main walls and vaults, the restoration of beams and entire roofs...]\\[1pt]
\textbf{Analysis} Both annotators saw that the court is citing the \emph{contract} rather than applying the \emph{statute}. The lease mirrors the language of article 606, which is a standard practice in commercial leases. However, the court is interpreting a contractual clause, not applying the Civil Code article. This is a type of model failure when there is an almost perfect lexical match that is deceptive (detailed in~\S\ref{app:error_analysis}).

\section{Quantitative Error Analysis}
\label{app:error_analysis}

To complement the qualitative analysis of~\S\ref{sec:qualitative}, we manually evaluated the 66 false positives produced by the supervised ensemble. Table~\ref{tab:fp_modes} shows the distribution. We focus on false positives because their failure modes leave observable traces in the chunk. False negatives, on the other hand, would require reasoning about absent evidence.

\begin{table}[h]
\centering
\footnotesize
\setlength{\tabcolsep}{3pt}
\begin{tabular}{lrrrr}
\toprule
\textbf{Failure mode} & \textbf{N} & \textbf{\%} & \textbf{Disag.} & \textbf{Conf.} \\
\midrule
\makecell[l]{Statutory language present,\\ \quad not applied by the court} & 37 & 56 & 28 & 0.75/0.76 \\
\makecell[l]{Right legal domain,\\ \quad wrong rule} & 28 & 42 & 17 & 0.76/0.76 \\
Other & 1 & 2 & 0 & 0.71/0.71 \\
\midrule
Total & 66 & 100 & 45 & 0.75/0.76 \\
\bottomrule
\end{tabular}
\caption{Failure modes of the supervised ensemble on its 66 FP. Confidence shown as mean/median.}
\label{tab:fp_modes}
\end{table}

The first category corresponds to cases where statutory language is used in the chunk but not applied by the court. Often, the article's vocabulary is echoed in a party's claim or a contract clause. The model mistakes the presence of legal language for its application (cf.\ the art.~2251 and art.~606 examples in Appendix~\ref{app:examples}). The second category of errors involves cases where the legal domain is the correct one but the rule retrieved is not the appropriate one. It can be a sibling article of the same domain or a special regime that displaces the general article (cf.\ the art.~1361 example in~\S\ref{sec:qualitative}). One case fits neither pattern. On 45 of the 66 pairs, the annotators disagreed, which is consistent with~\S\ref{sec:fp_analysis}.

\end{document}